\documentclass{article}

\usepackage{arxiv}
\usepackage[numbers]{natbib}

\usepackage[utf8]{inputenc} 
\usepackage[T1]{fontenc}    

\usepackage{amsmath,amssymb,amsfonts}
\usepackage{algorithmic}
\usepackage{graphicx}

\usepackage{verbatim}

\usepackage{enumitem}

\usepackage[hyphens]{url}
\usepackage{booktabs}
\usepackage{multirow}
\usepackage{cleveref}

\usepackage{tikz}
\usepackage{pgfplots}
\pgfplotsset{compat=1.16}
\usepgfplotslibrary{fillbetween}
\usetikzlibrary{positioning,fit,calc,arrows,shapes,backgrounds,plotmarks}
\pgfdeclarelayer{background}
\pgfsetlayers{background,main}
\makeatother
\newsavebox{\mybox}

\title{Design of Planar Collision-free Trochoidal Paths for a Multi-robot Swarm}

\author{
  Adil Shiyas \\
  International Institute of Information Technology Bangalore \\
  Bangalore, India\\
  \texttt{adil.shiyas@iiitb.ac.in} \\
   \And
  Sachit Rao \\
  International Institute of Information Technology Bangalore \\
  Bangalore, India\\
  \texttt{sachit@iiitb.ac.in}
}

\renewcommand{\shorttitle}{Trochoidal Paths for a Multi-robot Swarm}

\begin{document}
\maketitle

\begin{abstract}  
In the literature, a distributed consensus protocol by which a connected swarm of agents can generate artistic patterns in 2-dimensional space is proposed. Motivated by this protocol, in this paper, we design the parameters of this protocol for a 3-agent swarm of non-holonomic robots of finite size that results in the generation of periodic trochoidal trajectories that satisfy a set of geometric and speed constraints; this design also includes selecting the initial positions of the robots. This problem finds applications in persistent surveillance and coverage, guarding a region of interest, and target detection. While the trajectories may be self-intersecting, imposing geometric constraints \textit{i}. eliminates collisions between robots; \textit{ii}. ensures minimum and maximum separation distance between any robot and a fixed point, thus ensuring the robots are in communication range. Imposing speed constraints ensure that tracking these trajectories becomes feasible. It is also shown that robots can be injected to these paths at specific locations, in order to increase the \textit{refresh rate}, without violating any of the geometric constraints. The designs are implemented in an indoor mobile robot platform.
\end{abstract}

\keywords{Swarm Robotics; Trochoidal Patterns; Persistent Coverage; Surveillance}


\section{Introduction}

The distributed consensus protocol (CP) presented in \cite{Tsiotras2014} enables a connected swarm of agents, modelled as single integrators, to trace repeated geometric paths in both 2-dimensional (2-D) and 3-D spaces. The parameters of the protocol, the connection topology of the swarm, and the initial positions of the agents define the characteristics of the generated paths. The applications of agents, or robots, tracing such paths have been well identified in the literature, for example, persistent coverage and surveillance of a region, guarding an asset, and target detection; the cited references provide an exhaustive list. In such applications, the issues of collisions between robots, communication range, feasibility of path tracking, and time taken to trace the path, have to be considered explicitly. Further, in applications involving guarding a specific region or an asset, the path should be defined by making the asset the centre of rotation (CoR).

In this paper, the protocol is designed for a connected swarm of 3 unicycle-robots of finite size moving in a 2-D Cartesian space; the use of 3 robots leads to the generation of trochoidal paths, where each robot traces a unique trochoid. For a given communication topology, design of the trochoidal paths involves the selection of 3 scalars for the CP, a positive integer, $k$, with magnitude $k\geq2$, and the initial coordinates of the 3 robots in $X-Y$ space, thus leading to a total of 10 design variables. The 3 scalars in the CP and $k$ determine the nature of the trochoid - hypotrochoid or epitrochoid - and the number of cusps, respectively; it is remarked that for a chosen set of these variables, all robots trace the same type of trochoid. In addition to these variables, the initial positions influence the coordinates of the common CoR, distances between robots, and the closest and farthest distances of the robots from the CoR. Thus, the constraints of no collisions or maintaining a communication range can be ensured for an appropriate choice of the design variables. 

By analysing the properties of the CP and the common features of the resulting trochoids, the exercise of designing 10 variables is reduced to designing 4 variables: \textit{i}. one scalar that determines the number of cusps ($k)$; \textit{ii}. one set of positive integers that form a Pythagorean triple; and \textit{iii}. a pair of distance metrics, denoted by $(R_c,d_c)$. Since the trochoids have to satisfy numerous constraints, first, these constraints are expressed in the form of linear inequalities in terms of $(R_c,d_c)$ and second, for a choice of $k$ and a Pythagorean triple, regions bounded by these inequalities in the $(R_c,d_c)$ space are identified. Finally, any $(R_c,d_c)$ pair that lies in these feasible regions is selected, which in turn, decides the values of the original design variables. By selecting a design variable as a Pythagorean triple, the trochoids become closed, \cite{Fedele2023}, that is, the robots return to their respective initial positions after one period.

The use of trochoidal paths, for diverse applications, has received attention in the literature. In \cite{Fedele2023}, a consensus-based control policy - and a single scalar - is proposed by which a connected network of agents defined as double integrators, all trace trochoidal trajectories with the same parameters in the 2-D space; the choice of initial conditions ensures collision avoidance. This solution is extended to the 3-D case in \cite{Fedele2023TAC}, but, collision avoidance is not explicitly considered in the analysis. In \cite{MONSINGH201984} and \cite{M2023100928}, the eigenvalues of a design matrix are evaluated so that agents modeled as single and double integrators, respectively, trace trochoidal paths; even in these references, collision avoidance between agents is not explicitly considered. The CP adopted in this paper guarantees eigenvalues with the desired properties; their selection is made simpler still with the use of Pythagorean triples. In \cite{hegde2021}, a feedback control law is designed using Barrier Lyapunov Functions that allow a swarm of agents, moving at constant speed, to trace closed curves, such as trochoids (also the same for all agents); the control law balances the curve-phases of the agents in the complex plane. A cyclic pursuit based control policy is proposed in \cite{Pavone2007}, where the agents can trace either circular or logarithmic spiral patterns, based on the choice of a design parameter. A variant to trochoidal paths is the tracing of Lissajous curves, as described in \cite{Borkar2020,Boldrer2023}; by selecting the initial locations of the agents and the parameters of the Lissajous curve (which are naturally smooth), the agents guarantee sensor coverage of a rectangular region without colliding with each other. A guidance law is proposed in \cite{Ratnoo2019}, whose design parameter leads to a robot generating a trochoidal trajectory around a stationary target; the trochoidal trajectory is designed to maximise observability of the target based on range measurements. Bifurcation is used as the idea to generate limit cycles that are trochoidal patterns in \cite{Parayil2019}; in both \cite{Ratnoo2019} and \cite{Parayil2019}, a single unicycle robot is considered. In this paper, the 3 agents trace different trajectories, albeit of the same type, that is, hypotrochoidal or epitrochoidal; further, these results form a design methodology, along with extensions, to the protocol proposed in \cite{Tsiotras2014}.

From an implementation perspective, it is known that the properties of the trochoidal paths are closely tied to the initial positions. Thus, if the robots' initial positions are slightly perturbed, the resulting paths will be different and may also violate a few of the distance-based constraints. In this paper, a bound on such perturbations is identified so that the perturbed paths continue to remain feasible. It is shown that this can be achieved by choosing a feasible $(R_c,d_c)$ pair from ``large''  regions in the $R_c-d_c$ space. It is also highlighted that since analytical forms of the paths are known, these can be programmed in each robot, thus relaxing the need for communication, as now the robots can implement individual path tracking controllers to trace them. It is also shown that if need be, by scaling down the Pythagorean triples equally, speed and turn rate constraints that are present in hardware can be respected. Though the design is performed for a 3-robot swarm, it is also shown in this paper that additional robots can be injected at different locations on each trochoidal path, while ensuring collision avoidance with any other robot. With the introduction of these additional robots, the refresh period - the time taken by a single robot to trace the complete path, \cite{Keller2017} - is considerably reduced; the additional robots injected in one path follow the robot that traces this path by implementing the CP. 

The main contributions of the paper are as follows: the design of a CP for a 3-robot swarm that leads to them tracing closed trochoidal paths around a region of interest; the designed paths \textit{i}. ensure geometric and speed constraints are respected; \textit{ii}. guarantee collision avoidance; \textit{iii}. offer flexibility in implementation from a consensus-based approach to that of control by individual robots; \textit{iv}. allow for scaling of linear and angular speeds of the non-holonomic robots without causing any change in the resulting geometric patterns; and \textit{v}. can be traced with an increased refresh rate by the injection of additional robots. The placement of eigenvalues The proposed designs are also implemented in an indoor mobile robot platform to demonstrate the feasibility of the design and its robustness to perturbations in initial positions of the robots.

The paper is organised as follows: In Sec.~\ref{Sec:Prelim}, the robot characteristics and their communication topology, the CP presented in \cite{Tsiotras2014}, and the properties of the resulting trochoids are briefly described. In Sec.~\ref{Sec:DesTr}, the design of the CP parameters and the initial coordinates of the 3 robots is presented. In Sec.~\ref{Sec:Res}, simulation and experimental results of implementation are discussed. Concluding remarks are made in Sec.~\ref{Sec:Conc}.

\section{Preliminaries}\label{Sec:Prelim}

\subsection{Characteristics of the robots and the CP}\label{Sec:RobCPProp}

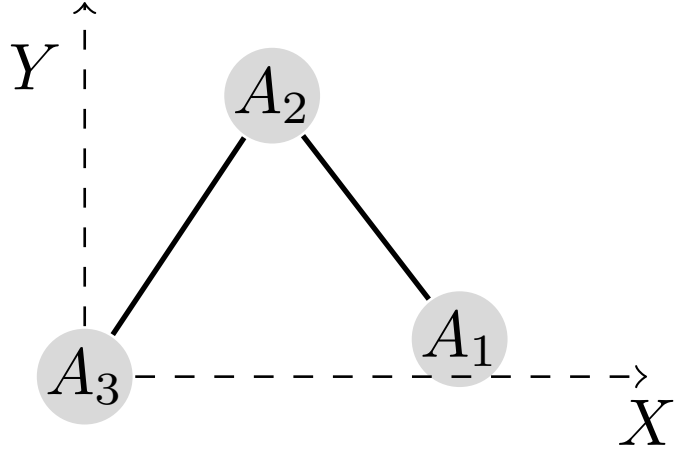
\begin{figure}[tpb!]
	\centering
	\resizebox{0.6\columnwidth}{!}{
		
		\begin{tikzpicture}
			
			\node[fill=gray!30,thick,circle,inner sep=0pt, minimum size = 0.5cm] (A3) at (0, 0) {$A_3$};
			\node[fill=gray!30,thick,circle,inner sep=0pt, minimum size = 0.5cm] (A2) at (1, 1.5) {$A_2$};
			\node[fill=gray!30,thick,circle,inner sep=0pt, minimum size = 0.5cm] (A1) at (2, 0.2) {$A_1$};
			
			\draw[style=thick] (A1) -- (A2); 
			\draw[style=thick] (A2) -- (A3); 
		
			\draw[->,dashed] (A3) -- (3,0) node[pos=1,below] {$X$};
			\draw[->,dashed] (A3) -- (0,2) node[pos=0.8,left] {$Y$}; 
			
		\end{tikzpicture}
		
		}
		\caption{A connected swarm of 3 agents moving in the plane} 
		\label{fig:3AgentFig}
\end{figure}

Consider a connected swarm of 3 homogeneous agents such as shown in Fig.~\ref{fig:3AgentFig}. The motion of the agents in the Cartesian $X-Y$ plane is defined by the dynamics $\dot{x}_i=u_{xi}, \ \dot{y}_i=u_{yi}, \ i=1,2,3$, where $(x_i,y_i)$ are the coordinates of agent $A_i$ in the Cartesian plane and $u_{xi},u_{yi}$ are the inputs to $A_i$. As discussed in \cite{Tsiotras2014}, the CP
\begin{equation}\label{CPExp}
    \mathbf{u}=\left(\left(\mathbf{BL}\right)\otimes\mathbf{S}\right)\mathbf{x}, \ \mathbf{u}=\left[u_{x1} \ u_{y1}\cdots\right]^T, \mathbf{x}=\left[x_1 \ y_1\cdots\right]^T,
\end{equation}
leads to the agents tracing trochoidal paths on the $X-Y$ plane; these paths have the analytical expressions
\begin{align}\label{TrajExp}
    \mathbf{x}(t) &= \mathbf{V}_L\left(\cos{\left(\mathbf{J}t\right)}\otimes\mathbf{I}_2 + \sin{\left(\mathbf{J}t\right)}\otimes\mathbf{S}\right)\mathbf{V}_R \mathbf{x}_0, \\
    \mathbf{V}_L &= \mathbf{V}\otimes\mathbf{I}_2, \ \mathbf{V}_R = \mathbf{V}^{-1}\otimes\mathbf{I}_2.\nonumber
\end{align}
In the CP \eqref{CPExp} and \eqref{TrajExp}, $\mathbf{B}\in\Re^{3\times3}$ is a user-defined diagonal matrix, $\mathbf{L}\in\Re^{3\times3}$ is the graph Laplacian, $\mathbf{S}$ is the 2-dimensional skew-symmetric matrix; $\mathbf{x}_0$ are the agents' initial coordinates; $(\mathbf{V},\mathbf{J})$ are the eigenvector and eigenvalue matrices, respectively, of $\left(\mathbf{BL}\right)$; and $\mathbf{I}_2$ is the 2-D identity matrix. The operator $\otimes$ denotes the Kronecker product.

Suppose the agents are differential drive robots with non-holonmic constraints, then these can be modeled as unicycles in the form
\begin{equation}\label{AgentKin}
    \dot{x}_i = V_i\cos{\gamma_i}, \ \dot{y}_i = V_i\sin{\gamma_i}, \ \dot{\gamma}_i =  \omega_i, 
\end{equation}
where $(x,y,\gamma)_i$ denotes the pose of robot $A_i$, $|V_i|\leq V_{\max}$ is its linear speed, and $|\omega_i|\leq \omega_{\max}$ is the angular speed. Now, the CP \eqref{CPExp} can be implemented by calculating the linear speed $V_i=\left(u_{xi}^2+u_{yi}^2\right)^{0.5}$, the desired orientation $\gamma_{i\text{Ref}}=\tan^{-1}\left(u_{yi}/u_{xi}\right)$, and the angular speed as the proportional controller $\omega_i=K_P\left(\gamma_{i\text{Ref}}-\gamma_{i}\right)$ with gain $K_P>0$; note that $\omega_i$ can also be designed as a proportional-integral controller.

\subsection{Trochoids' properties}

The type of trochoid is defined by the ratio of the largest and smallest eigenvalues of $\left(\mathbf{BL}\right)$; note that one of the eigenvalues is always zero. Now, by defining $\mathbf{B}=\text{diag}\left(\left[\beta_1 \ \beta_2 \ \beta_3\right]\right)$, the other two eigenvalues are given by
\begin{equation}\label{EigBLVal}
    \lambda_{1,2} = \frac{\beta_1+2\beta_2+\beta_3}{2} \pm \frac{1}{2}\left(\left(\beta_1-\beta_3\right)^2 + \left(2\beta_2\right)^2\right)^{0.5}.
\end{equation}
It is the 3 scalars $\beta_i$ that form the key design variables of the CP. Suppose these are chosen such that the two eigenvalues, denoted by $\lambda_{\max}>\lambda_{\min}$, are distinct. Now, if the ratio $\frac{\lambda_{\max}}{\lambda_{\min}}$ is an \textit{integer}, then the trochoidal patterns are closed. Further, by introducing the integer design variable $k\geq2$, if $\frac{\lambda_{\max}}{\lambda_{\min}} = +(k+1)$, then the path becomes an epitrochoid; if $\frac{\lambda_{\max}}{\lambda_{\min}} = -(k-1)$, then the path becomes a hypotrochoid. It is clear that the integer $k$ represents the number of cusps in the corresponding trochoid.

In terms of these eigenvalues, the expressions \eqref{TrajExp} for agent $A_i$ have the form
\begin{subequations}\label{XYlmaxminExp}
    \begin{align}
        x_i(t) &= c_{ir}\cos{\left(\lambda_{\min}t + \phi_{r}\right)} + c_{id}\cos{\left(\lambda_{\max}t + \phi_{d}\right)}  + c_{i0}\cos{\phi_{0}} \label{xlmaxminExp} \\ 
        y_i(t) &= c_{ir}\sin{\left(\lambda_{\min}t + \phi_{r}\right)} + c_{id}\sin{\left(\lambda_{\max}t + \phi_{d}\right)} + c_{i0}\sin{\phi_{0}},\label{ylmaxminExp}    
    \end{align}
\end{subequations}
where the coefficients $c_{ij}$ and the phase angles $\phi_{j}, \ j=r,d,0$, are given by
\begin{align}\label{cijphiijExp}
    c_{ij} &= \mathbf{V}_{i,p} \left(q_x^2 + q_y^2\right)^{0.5}, \ \phi_{j} = \tan^{-1}\left(\frac{q_y}{q_x}\right) \\
    q_x &= \begin{bmatrix} \mathbf{V^{-1}}_{p,1} & \mathbf{V^{-1}}_{p,2} & \mathbf{V^{-1}}_{p,3} \end{bmatrix} \begin{bmatrix} x_{10} \\ x_{20} \\ x_{30} \end{bmatrix}, \ q_y = \begin{bmatrix} \mathbf{V^{-1}}_{p,1} & \mathbf{V^{-1}}_{p,2} & \mathbf{V^{-1}}_{p,3} \end{bmatrix} \begin{bmatrix} y_{10} \\ y_{20} \\ y_{30} \end{bmatrix} \nonumber
\end{align}
The indices $j$ and $p$ are given by
\begin{equation}\label{pjindDef}
    p = \begin{cases}
        \text{ index of } \lambda_{min} \text{ of } \mathbf{J} & \text{for } j=r \\
        \text{ index of } \lambda_{max} \text{ of } \mathbf{J} & \text{for } j=d \\
        \text{ index of the \textbf{zero} eigenvalue}  \text{ of } \mathbf{J} & \text{for } j=0
        \end{cases}
\end{equation}
Note that as $\mathbf{J}$ is the diagonal matrix of eigenvalues, the index of any eigenvalue is given by its location $(p,p)$.

The eigenvector corresponding to the zero eigenvalue of $\left(\mathbf{BL}\right)$ consists of identical elements. Hence, the coefficients $c_{i0}=c_{j0}=c_0, \ i,j=1,2,3, \ i\neq j$. This implies that all agents trace trochoidal patterns about the point with coordinates $(c_{0}\cos{\phi_{0}},c_{0}\sin{\phi_{0}})$ as the centre. Now, by placing the agents at the ``new'' initial positions $x_{i0}'=x_{i0}-c_{0}\cos{\phi_{0}}$ and $y_{i0}'=y_{i0}-c_{0}\sin{\phi_{0}}$, the CoR can be made to be the origin of the $X-Y$ plane.

With this background, the main problems that are addressed in this paper are: Design the \textit{i}. elements of the matrix $\mathbf{B}$, \textit{ii}. the number of cusps $k$, and \textit{iii}. the initial positions, to obtain closed and periodic epi- or hypotrochoidal paths, such that
\begin{enumerate}
    \item the closest, and farthest, distance to the origin of any path is larger, or smaller, than a user-defined metric;
    \item $\forall \ t\geq0$, the robots are separated by some distance, say as defined by the size of the agents; and
    \item the speed and turn rate constraints are not violated.
\end{enumerate}
These objectives ensure that the robots do not cross into the guarded region, in such applications, and trace large arcs that may impose communication challenges if they are equipped with limited range sensors, and also that they do not collide with each other while tracking the paths. 

\section{Design of the Trochoids}\label{Sec:DesTr}

The design of trochoids is demonstrated for an epitrochoid; hypotrochoids can be designed similarly. As mentioned in the Introduction, to show that designing 10 variables - the parameters of the CP, $\beta_{1,2,3}$, the number of cusps $k$, and the initial coordinates $\mathbf{x}_0$ - can be reduced to designing just 4, the coefficients $c_{ir},c_{id}$ in \eqref{XYlmaxminExp} are analysed. It can be shown that these coefficients can be expressed in the form 
\begin{equation}\label{ceqs}
    c_{ir} = \alpha_{ir}R_c, \ c_{id} = \alpha_{id}d_c,
\end{equation}
where $\alpha_{ir}$ and $\alpha_{id}$ are controlled solely by $\mathbf{B}$ while $R_c$ and $d_c$ are influenced by $\mathbf{B}$ as well as $\mathbf{x}_0$. These are given by
\begin{align}\label{AlphaVal}
    \alpha_{1r} &= \frac{1}{\beta_d} \frac{\beta_1}{\beta_2}\left(\lambda_{\min}-\beta_2-\beta_3\right), \ \alpha_{1d} = \frac{1}{\beta_d} \frac{\beta_1}{\beta_2}\left(\lambda_{\max}-\beta_2-\beta_3\right), \\
    \alpha_{2r} &= \frac{\beta_3 - \lambda_{\min}}{\beta_d}, \ \alpha_{2d} = \frac{\beta_3 - \lambda_{\max}}{\beta_d}, \ \alpha_{3r} = \alpha_{3d} = \frac{\beta_3}{\beta_{d}} \nonumber
    \intertext{where}
    \beta_{d} &= 4b\left(\beta_1\beta_2 + \beta_2\beta_3+\beta_1\beta_3\right), \nonumber\\
    a &= \frac{\beta_1}{2} + \beta_2 + \frac{\beta_3}{2}, \ b = \frac{1}{2}\left(\left(\beta_1-\beta_3\right)^2 + \left(2\beta_2\right)^2\right)^{0.5},\nonumber
\end{align}
and
\begin{subequations}\label{rcdcEqs}
    \begin{align}
        R_c &= \left( \left( \Gamma_R^T\mathbf{x}_{i0}\right)^2 + \left( \Gamma_R^T\mathbf{y}_{i0} \right)^2\right)^{0.5}, \ d_c = \left(\left( \Gamma_d^T\mathbf{x}_{i0}\right)^2 + \left( \Gamma_d^T\mathbf{y}_{i0} \right)^2\right)^{0.5}, \label{RcdcEq}\\
        \phi_r &= \tan^{-1}\left( \frac{ \Gamma_{\phi r}^T\mathbf{y}_{i0} }{\Gamma_{\phi r}^T\mathbf{x}_{i0}} \right), \ \phi_d = \tan^{-1}\left( \frac{\Gamma_{\phi d}^T\mathbf{y}_{i0} }{\Gamma_{\phi d}^T\mathbf{x}_{i0} }\right) \label{phirdEq} 
    \end{align}
\end{subequations}
where $\mathbf{x}_{i0}=[x_{10} \ x_{20} \ x_{30}]^T, \ \mathbf{y}_{i0}=[y_{10} \ y_{20} \ y_{30}]^T$, and the coefficient vectors $\Gamma_R,\Gamma_d,\Gamma_{\phi r}, \Gamma_{\phi d}$ are defined in the Appendix.

\subsection{Design of $\mathbf{B}$}

For a closed epitrochoid with a user-defined number of cusps $k$, the ratio of the non-zero eigenvalues of $\mathbf{BL}$, given by \eqref{EigBLVal}, should also be an integer and satisfy $\frac{\lambda_{\max}}{\lambda_{\min}} = +(k+1)$. The parameters $\beta_i$ are selected such that they form a Pythagorean triple, defined by $(s_1,s_2,s_3)$, where $s_3^2=s_1^2+s_2^2$. With this choice, these parameters are given by
\begin{equation}\label{BValss123}
    \beta_1 = \frac{2s_3 - k\left(s_2 + s_1 - s_3 \right)}{2k}, \ \beta_2 = \frac{s_2}{2}, \ \beta_3 = \beta_1 + s_1.
\end{equation}
It can be seen that for a specific triple $(s_1,s_2,s_3)$ and some $k$, the equality $\beta_1=0$ may hold, for example, for $k=5$ and the triple $(3,4,5)$, including its multiples (that makes them real numbers). In such cases, from \eqref{AlphaVal}, it can be seen that the coefficients $\alpha_{1r}=\alpha_{1d}=0$, implying that agent $A_1$ does not move at all. This implies that for agents $A_2$ and $A_3$, their CoR is agent $A_1$; such choices of $(s_1,s_2,s_3)$ and $k$ should be avoided. Note that for hypotrochoids, the parameter $\beta_1=\frac{-2s_3 - k\left(s_2 + s_1 - s_3 \right)}{2k}$, while $\beta_2,\beta_3$ are the same as in \eqref{BValss123}.
                        
The benefit offered with the use of a Pythagorean triple, is that by scaling up, or down, the triple $(s_1,s_2,s_3)$, the coefficients $c_{ir},c_{id}$, and hence, the magnitudes of the derivatives of $x_i(t),y_i(t)$, are also correspondingly scaled. This feature proves helpful in hardware implementation on robots that have bounded linear and angular speed capabilities.

\subsection{Design of the Initial Positions}\label{Sec:InitPos}

The CP \eqref{CPExp}, when expanded, reveals a form that is dependent on the \textit{difference} in the agents' positions, rather than their absolute values. Thus, choosing the agents' initial positions to design the trochoidal trajectories can be simplified by placing one of them at the origin and finding the positions of others relative to this origin. While this approach leads to a CoR that is not the origin, once the trochoid is designed, all agents can again be shifted by known distances so that the trochoids are centered at the origin or some desired location. As a result, without any loss of generality, let agent $A_3$ be placed at the origin; this choice makes $(x_{30}, y_{30}) = (0,0)$.

To simplify further choosing initial positions, consider the distance of agent $A_i$ from the origin, given by $\sqrt{x_i(t)^2+y_i(t)^2}$. It can be shown that, for $\alpha_{ir},\alpha_{id}>0$, $A_i$ is closest to the origin at
\begin{equation*}
    t_{\min} = \frac{1}{k\lambda_{min}} \left( 2(m+1)\pi - \left( \phi_d - \phi_r\right)\right), \ m=0,1,\cdots,
\end{equation*}
and farthest at the time instants
\begin{equation*}
    t_{\max} = \frac{1}{k\lambda_{min}} \left( 2m\pi - \left( \phi_d - \phi_r\right)\right), \ m=0,1,\cdots.
\end{equation*}
Note that if $\alpha_{ir},\alpha_{id}$ have different signs, then $t_{\min}$ is when the agent is farthest and $t_{\max}$ is when it is the closest. 

It is clear from these expressions that the difference $\left( \phi_d - \phi_r\right)$ is common to \textit{all} agents and essentially offsets, in time, when an agent is closest to, or farthest from, the origin. A similar analysis holds for pair-wise distances between the agents; the term $\left( \phi_d - \phi_r\right)$ shifts when a pair of agents are closest to, or farthest from, each other. Thus, without any loss of generality, this offset can be set to zero, that is, the initial positions can be selected so that $\phi_r=\phi_d$, in \eqref{rcdcEqs}. With this choice, it can be shown that the agents are now co-linear and lie on a line with slope $\tan{\phi_r}$. Further, since the actual magnitudes of $\phi_r,\phi_d$ do not influence the epitrochoidal paths, for simplicity, they can be set to zero. From \eqref{phirdEq}, this implies that the agents can all be selected to lie on the $X-$axis of the global Cartesian plane; note that this is a special case of the line on which the agents are co-linear. Thus, for the 3-agent swarm, the $x-$coordinates of agents 1 and 2, $x_{10}$ and $x_{20}$, respectively, become the design variables. 

To satisfy the stated geometrical constraints, which are in essence distance measures, from \eqref{rcdcEqs}, the design of $x_{10,20}$, is transformed to finding the appropriate values of $R_c,d_c$. This is possible since the coefficients $\Gamma_R,\cdots$ have already been found, implying that $\alpha_{ir},\alpha_{id}$ are also known. Thus, the distance constraints can be satisfied by choosing $R_c,d_c$.

\subsection{Design of $R_c,d_c$}\label{Sec:RcdcDes}

The parameters $R_c,d_c$ are determined by points in the $R_c-d_c$ space that are bounded by a set of inequalities. These inequalities capture the constraints that
\begin{enumerate}
    \item the closest distance between the origin and any agent, denoted by $d_{i0\min}$, satisfies $d_{i0\min}\geq d_{0\min}>0$;
    \item the farthest distance from the origin to any agent, denoted by $d_{i0\max}$, satisfies $d_{i0\max}\leq d_{0\max}$;
    \item the closest distance between any pair of agents, $A_i,A_j$, given by $d_{ij\min}$, satisfies $d_{ij\min}\geq d_{CT}>0$; and
    \item the farthest distance between any pair of agents, $A_i,A_j$, given by $d_{ij\max}$, satisfies $d_{ij\max}\leq d_{CR}$.
\end{enumerate}
The distance $d_{CT}$ can be interpreted as a collision threshold, which should not be crossed by a pair of agents, while the distance $d_{CR}$ can be interpreted as a limit on the communication range between a pair of agents. The value for $d_{CT}$ can be selected by modeling the agents as being bounded in size by a circle of radius $R_{\text{Rob}}$, thus $d_{CT}>2R_{\text{Rob}}$. Similarly, the distance $d_{0\min}$ can also be selected by modeling the asset, around which the agents traverse, as a circle of radius, say $R_{\text{Asset}}$, thus, $d_{0\min}>R_{\text{Asset}}$. Also note that if the communication range $d_{CR}$ is known, then the distance $d_{0\max}$ can be approximated as $d_{0\max}\approx0.5d_{CR}$.

In terms of $R_c,d_c$, for agent $A_i, \ i=1,2,3$, it can be shown that these constraints are defined by the inequalities that are linear in $R_c,d_c$ and given by
\begin{subequations}\label{IneqExp}
    \begin{align}
        & \big||\alpha_{ir}|R_c  -  |\alpha_{id}|d_c\big|\geq d_{0\min} \label{CloseDistIneq}\\
        & \big||\alpha_{ir}|R_c + |\alpha_{id}|d_c\big|\leq d_{0\max}  \label{FarDistIneq}\\
        & \big||\alpha_{ir}-\alpha_{jr}|R_c - |\alpha_{id}-\alpha_{jd}|d_c\big|\geq d_{CT}, \ i\neq j \label{PairCloseDistIneq}\\
        & \big||\alpha_{ir}-\alpha_{jr}|R_c + |\alpha_{id}-\alpha_{jd}|d_c\big|\leq d_{CR}, \ i\neq j \label{PairFarDistIneq} \\
        & R_c\geq0, \ d_c\geq0 \label{RcdcIneg}
    \end{align}
\end{subequations}
For the 3-agent swarm, from a geometrical perspective, the values of $\mathbf{B}$ and $(R_c,d_c)$ that satisfy these 14 inequalities lead to a feasible set of epitrochoidal paths. As is evident, these inequalities form polyhedra in the $R_c-d_c$ space. 

\begin{figure*}[!tbp]
  \centering
    \begin{lrbox}{\mybox}%

    \begin{tikzpicture}
        \begin{axis}[scale only axis, axis equal, xlabel= $R_c$, ylabel = $d_c$]
            \addplot[name path = t1, line width=2pt, color = blue] table {
            1449.5    929.2
            1574.4    1054.0
            2379.0    1657.5
            2539.5    1256.2
            } -- cycle;
            \addplot[name path = t2, line width=2pt, color = blue] table {
            343.5    2420.5
                 0    1647.8
                 0    2535.0
            } -- cycle;
            \addplot[name path = t3, line width=2pt, color = blue] table {
            2892.6    373.5
            3042.0         0
            1647.7         0
            } -- cycle;
            \addplot[thick, color = blue, fill = blue, fill opacity = 0.5]fill between[of=t1 and t2];
            \addplot[thick, color = blue, fill = blue, fill opacity = 0.5]fill between[of=t2 and t3];
            \addplot[thick, color = blue, fill = blue, fill opacity = 0.5]fill between[of=t3 and t1];
            \node[circle,fill,inner sep=3pt,black] at (axis cs:2500, 0) {};
            \node[circle,fill,inner sep=3pt,green] at (axis cs:2551.2, 159.277) {};
            \node[circle,fill,inner sep=3pt,red] at (axis cs:2000, 1200) {};

            \addplot[name path = l1, line width=2pt, color = blue] table {
            0   0
            1000    750
            2000    1500
            3000    2250
            };
            \addplot[name path = l2, line width=2pt, color = red] table {
            0   0
            1000    100
            2000    200
            3000    300
            };
            \addplot[name path = l3, line width=2pt, color = green] table {
            0   0
            1000    333.33
            2000    666.66
            3000    1000
            };
            
        \end{axis}
    \end{tikzpicture}
    
    \hfill
    \begin{tikzpicture}
    \begin{axis}[axis equal, scale only axis, xlabel= $X$, ylabel = $Y$]
        \addplot[name path = A1, color = blue] table{A1_2500_0_C2.txt};
        \addplot[name path = A2, color = red] table{A2_2500_0_C2.txt};
        \addplot[name path = A3, color = green] table{A3_2500_0_C2.txt};
        \addplot[name path = A1_E, color = blue, dashed] table{CIRC_ERROR_A1.txt};
        \addplot[name path = A2_E, color = red, dashed] table{CIRC_ERROR_A2.txt};
        \addplot[name path = A3_E, color = green, dashed] table{CIRC_ERROR_A3.txt};
        \node[circle,fill,inner sep=2.5pt,blue] at (axis cs:-5.1210000000000004e+00, 0.0000000000000000e+00) {};
        \node[circle,fill,inner sep=2.5pt,red] at (axis cs:2.2759999999999998e+00, 0.0000000000000000e+00) {};
        \node[circle,fill,inner sep=2.5pt,green] at (axis cs:7.2069999999999999e+00, 0.0000000000000000e+00) {};
        \node[diamond,fill,inner sep=2pt,blue] at (axis cs:-5.3830, -0.4820) {};
        \node[diamond,fill,inner sep=2pt,red] at (axis cs:2.7770, -0.2140) {};
        \node[diamond,fill,inner sep=2pt,green] at (axis cs:6.8720, 0.1750) {};
        \node[circle,fill,inner sep=2pt,black] at (axis cs:0,0) {};
        \end{axis}
    \end{tikzpicture}%

    \hfill
    \begin{tikzpicture}
    \begin{axis}[axis equal, scale only axis, xlabel= $X$, ylabel = $Y$]
        \addplot[name path = A1, color = blue] table{A1_2000_1200_C2.txt};
        \addplot[name path = A2, color = red] table{A2_2000_1200_C2.txt};
        \addplot[name path = A3, color = green] table{A3_2000_1200_C2.txt};
        \node[circle,fill,inner sep=2.5pt,blue] at (axis cs:-3.0040, 0) {};
        \node[circle,fill,inner sep=2.5pt,red] at (axis cs:-1.8210, 0) {};
        \node[circle,fill,inner sep=2.5pt,green] at (axis cs:9.2250, 0) {};
        \node[circle,fill,inner sep=2pt,black] at (axis cs:0,0) {};
        \end{axis}
    \end{tikzpicture}%
    
    \end{lrbox}
    \resizebox{0.95\textwidth}{!}{\usebox{\mybox}}
    \caption{Design parameters $k = 2$, $s_{1,2,3}=\left(5,12,13\right)$, and $d_{0\min}$ = 1.5, $d_{0\max} = d_{CR}= 15$, and $d_{CT} = 0.5$. \textit{Left}: Feasible regions in the $R_c-d_c$ space; \textit{Center}: Paths for $R_c=2500,d_c=0$; \textit{Right}: Paths for $R_c=2000,d_c=1200$. Blue is Agent $A_1$, red is Agent $A_2$, and green is Agent $A_3$. The straight lines shown in the left sub-figure represent $R_c,d_c$ values that lead to trochoidal paths that have cusps, which are locations where the angular speed of a non-holonomic robot is undefined.}
    \label{fig:DiffRcDcEx}
\end{figure*}
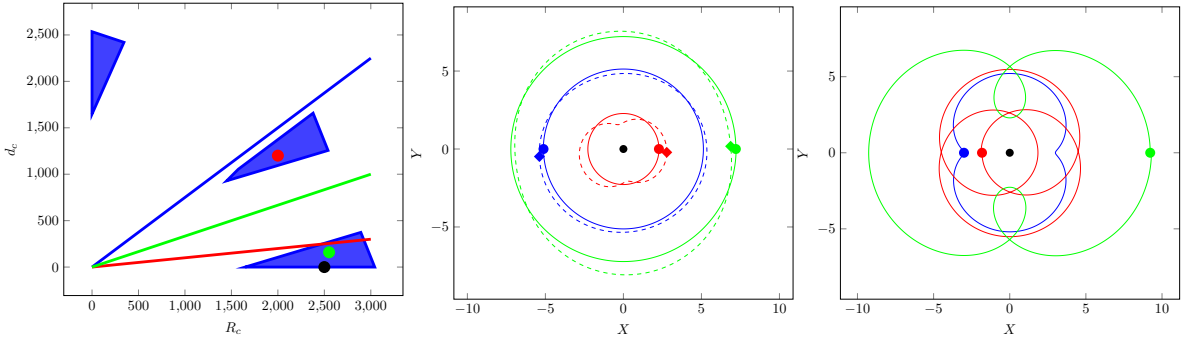

\textit{Example}: Choosing $k = 2$, the Pythagorean triple = $\left(5,12,13\right)$, and the distances $d_{0\min}$ = 1.5, $d_{0\max} = d_{CR}= 15$, and $d_{CT} = 0.5$ leads to the formation of the regions shown in the left sub-figure of Fig.~\ref{fig:DiffRcDcEx}. Selection of a particular $\left(R_c, d_c \right)$ pair from these regions determines the path. For example, for the solution $R_c=2500,d_c=0$ (marked by the black circle in the left sub-figure in Fig.~\ref{fig:DiffRcDcEx}), the paths become circles (denoted by the solid lines in the center sub-figure in Fig.~\ref{fig:DiffRcDcEx}), while for $R_c=2000,d_c=1200$ (marked by the red circle in the left sub-figure in Fig.~\ref{fig:DiffRcDcEx}), the paths are epitrochoids (also the solid lines in the right sub-figure in Fig.~\ref{fig:DiffRcDcEx}). For a feasible $(R_c,d_c)$ pair, the initial coordinates $x_{10},x_{20}$ are obtained from \eqref{RcdcEq}; note that, by construction, $x_{30}=0$ and $y_{i0}=0 \ \forall \ i$. In both the figures in Fig.~\ref{fig:DiffRcDcEx}, the agents' initial positions are recalculated to shift the CoR to the origin - as can be seen, this recalculation shifts only the $x-$coordinates of the initial positions.

\subsection{Injecting Additional Agents}

In this section, conditions are derived when additional agents can be injected to the 3 epitrochoidal paths designed following the described procedure. If agent $A_i,\ i=1,2,3,$ traces path $P_i$, then additional agents $A_{mi}, \ m=1,\cdots$, are injected to path $P_i$. It is clear that agents $A_{mi}$ should be placed such that they do not collide with other agents $A_{j}, \ j\neq i$ and those injected on paths $P_j$; note that agents on the same path, if injected at different locations, naturally do not collide with each other. Consider two epitrochoids in the parametric form
\begin{subequations}\label{XYEpiA_kA_j}
    \begin{align}
        x_{mi} &= \left(k+1\right)r_ic_{\left(\theta + \phi\right)} - d_ic_{\left(\left(k+1\right)\left(\theta + \phi\right)\right)} \label{xEpiAk} \\
        y_{mi} &= \left(k+1\right)r_is_{\left(\theta + \phi\right)} - d_is_{\left(\left(k+1\right)\left(\theta + \phi\right)\right)} \label{yEpiAk}
        \intertext{and}
        x_j &= \left(k+1\right)r_jc_{\left(\theta\right)} - d_jc_{\left(\left(k+1\right)\theta\right)} \label{xEpiAj} \\
        y_j &= \left(k+1\right)r_js_{\left(\theta\right)} - d_js_{\left(\left(k+1\right)\theta\right)} \label{yEpiAj},
    \end{align}
\end{subequations}
where $(r_i,d_i)$ and $(r_j,d_j)$ are the parameters of paths $P_i$ and $P_j$, respectively. In \eqref{xEpiAk} and \eqref{yEpiAk}, the angle $\phi$ defines the location of agent $A_{mi}$ that is offset from agent $A_i$ on the path $P_i$. 

Thus, for agents $A_{mi}$ and $A_{j}$ to avoid collisions, a suitable $\phi$ should be found. To avoid collisions, the squared separation distance between $A_{mi}$ and $A_{j}$, given by $d_{mij}^2 = (x_{mi}-x_j)^2 + (y_{mi}-y_j)^2$, should satisfy $d_{mij}^2\geq d_{CT}^2$, where $d_{CT}$ is the collision avoidance threshold. On expanding, it follows that $d_{mij}^2$ is a non-linear function of $\phi$ and hence to find the infeasible values of $\phi$, which are the solutions to the minimisation problem $d_{mij}^2- d_{CT}^2=0$, numerical routines may have to be adopted. However, for some specific values of $\phi$, a closed-form solution for the \textit{feasible} solutions can be obtained; this procedure is now discussed. 

Choose $\phi=\frac{\pi}{2}$ and let $k=2n, \ n=1,3,5,\cdots$. For this case, the difference $\delta_{mij}^2 = (d_{mij}^2 - d_{CT}^2)$ has the expression
\begin{align}\label{SqDistAkAjdCT_Keven}
     &(k+1)^2\left(r_i^2+r_j^2\right) + \left(d_i^2+d_j^2\right) \nonumber\\
     & + 2(k+1)\left(r_id_i - r_jd_j\right)\cos{\left(k\theta\right)} + 2(k+1) \left( r_id_j + r_jd_i\right)\sin{\left(k\theta\right)} - d_{CT}^2
\end{align}
and the values of $\theta$ when this difference has a minimum/maximum are given by $k\theta = \tan^{-1}{\left(\frac{r_id_j+r_jd_i}{r_id_i-r_jd_j}\right)}$ and $k\theta = \tan^{-1}{\left(\frac{r_id_j+r_jd_i}{r_id_i-r_jd_j}\right)} + \pi$. Substituting the first of these solutions in \eqref{SqDistAkAjdCT_Keven} results in the difference $\delta_{mij}^2$ achieving its minimum value, which is given by
\begin{equation}\label{Mindelij_Keven}
    \delta_{mij\min Ke}^2 = \left((k+1)\left(r_i^2+r_j^2\right)^{0.5} - \left(d_i^2+d_j^2\right)^{0.5} \right)^2 - d_{CT}^2 .
\end{equation}
Thus, if the trochoidal parameters, $r_{i,j}$, $d_{i,j}$, and $k$ are such that $\delta_{mij\min Ke}^2\geq0$, then agent $A_{mi}$ can be injected in the path of agent $A_i$, but offset by an angle $\phi=\frac{\pi}{2}$, so that it does not collide with agent $A_j$. For the case when $k=2n$, where $n=2,4,\cdots$, the result \eqref{Mindelij_Keven} also holds, but achieves its minimum at the second solution of $k\theta$. When $k$ is odd, the minimum of the difference $\delta_{mij}^2$ is given by
\begin{equation}\label{Mindelij_Kodd}
    \delta_{mij\min Ko}^2 = \left((k+1)\left(r_i^2+r_j^2\right)^{0.5} - \left(d_i+d_j\right) \right)^2 - d_{CT}^2.
\end{equation} 
In this case, the trochoidal parameters, $r_{i,j}$, $d_{i,j}$, and $k$ should be selected such that the inequality $\delta_{mij\min Ko}^2\geq0$ holds. It can be shown that the inequalities $\delta_{mij\min Ke}^2\geq0$ and $\delta_{mij\min Ko}^2\geq0$ should be satisfied for $\phi=\frac{3\pi}{2}$ as well.

For the case $\phi=\pi$, the expressions for the minimum of $\delta_{mij}^2$ is dependent on the signs of the products $\left(r_i+r_j\right)\left(d_i+d_j\right)$ and $\left(r_i+r_j\right)\left(d_j-d_i\right)$. It is remarked that, from the definitions of the coefficients in \eqref{ceqs} and comparing them with the coefficients of the parametric forms in \eqref{XYEpiA_kA_j}, the signs of these products are the same as that of the products $\left(\alpha_{ir}+\alpha_{jr}\right)\left(\alpha_{id}+\alpha_{jd}\right)$ and $\left(\alpha_{ir}+\alpha_{jr}\right)\left(\alpha_{jd}-\alpha_{id}\right)$, respectively. Thus, the following conditions can be derived for $\phi=\pi$:
\begin{enumerate}[wide = 0pt]
    \item For $k$ \textit{even} and $\left(\alpha_{ir}+\alpha_{jr}\right)\left(\alpha_{id}+\alpha_{jd}\right)>0$, the minimum distance, $\delta_{mij\min Ke}^2 = \left((k+1)\left(r_i+r_j\right) - \left(d_i+d_j\right) \right)^2 - d_{CT}^2$;
    \item For $k$ \textit{even} and $\left(\alpha_{ir}+\alpha_{jr}\right)\left(\alpha_{id}+\alpha_{jd}\right)<0$, $\delta_{mij\min Ke}^2 = \left((k+1)\left(r_i+r_j\right) + \left(d_i+d_j\right) \right)^2 - d_{CT}^2$;
    \item For $k$ \textit{odd} and $\left(\alpha_{ir}+\alpha_{jr}\right)\left(\alpha_{id}-\alpha_{jd}\right)>0$, $\delta_{mij\min Ko}^2 = \left((k+1)\left(r_i+r_j\right) - \left(d_j-d_i\right) \right)^2 - d_{CT}^2$;
    \item For $k$ \textit{odd} and $\left(\alpha_{ir}+\alpha_{jr}\right)\left(\alpha_{id}-\alpha_{jd}\right)<0$, $\delta_{mij\min Ko}^2 = \left((k+1)\left(r_i+r_j\right) + \left(d_j-d_i\right) \right)^2 - d_{CT}^2$;
\end{enumerate}
Thus, by considering the constraints that should hold for $\phi=\frac{\pi}{2},\pi,$ and $\frac{3\pi}{2}$, 3 additional agents can be injected in each path, increasing the number of agents to 12; the number of constraints also increases \textit{by} 18 (6 constraints for the 3 values of $\phi$ per path, that has to be checked for 3 paths). 

\begin{figure}[htpb!]
    \centering
    \begin{lrbox}{\mybox}%
    \begin{tikzpicture}
        \begin{axis}[scale only axis]
            \addplot[name path = t1, line width=2pt, color = blue] table {
            1449.5    929.2
            1574.4    1054.0
            2379.0    1657.5
            2539.5    1256.2
            } -- cycle;
            \addplot[name path = t2, line width=2pt, color = blue] table {
            343.5    2420.5
                 0    1647.8
                 0    2535.0
            } -- cycle;
            \addplot[name path = t3, line width=2pt, color = blue] table {
            2892.6    373.5
            3042.0         0
            1647.7         0
            } -- cycle;
            \addplot[name path = mt1, line width=2pt, color = red] table {
            2427.6    1535.9
            2288.8    1461.6
            2420.2    1554.5
            } -- cycle;
            \addplot[name path = mt2, line width=2pt, color = red] table {
            343.5    2420.5
            217.3    2136.6
            230.2    2458.3
            } -- cycle;
            \addplot[name path = mt3, line width=2pt, color = red] table {
            2892.6    373.5
            3042.0         0
            1647.7         0
            } -- cycle;
            \addplot[thick, color = blue, fill = blue, fill opacity = 0.5]fill between[of=t1 and t2];
            \addplot[thick, color = blue, fill = blue, fill opacity = 0.5]fill between[of=t2 and t3];
            \addplot[thick, color = blue, fill = blue, fill opacity = 0.5]fill between[of=t3 and t1];
            \addplot[thick, color = red, fill = red, fill opacity = 0.5]fill between[of=mt1 and mt2];
            \addplot[thick, color = red, fill = red, fill opacity = 0.5]fill between[of=mt2 and mt3];
            \addplot[thick, color = red, fill = red, fill opacity = 0.5]fill between[of=mt3 and mt1];
          \end{axis}
        \end{tikzpicture}%

        \hfill
        
        \begin{tikzpicture}
        \begin{axis}[axis equal, scale only axis]

        \addplot[name path = A1, color = blue] table{A1_2400_1525_C2.txt};
        \addplot[name path = A2, color = red] table{A2_2400_1525_C2.txt};
        \addplot[name path = A3, color = green] table{A3_2400_1525_C2.txt};
        
        \node[circle,fill,inner sep=2pt,blue] at (axis cs:-3.5280, 0) {};
        \node[circle,fill,inner sep=2pt,red] at (axis cs:-2.4430, 0) {};
        \node[circle,fill,inner sep=2pt,green] at (axis cs:11.3150, 0) {};
        
        \node[circle,fill,inner sep=2pt,blue] at (axis cs:0, -6.3040) {};
        \node[circle,fill,inner sep=2pt,red] at (axis cs:0, 6.8130) {};
        \node[circle,fill,inner sep=2pt,green] at (axis cs:0, 2.5230) {};
        
        \node[circle,fill,inner sep=2pt,blue] at (axis cs:3.5280, 0) {};
        \node[circle,fill,inner sep=2pt,red] at (axis cs:2.4430, 0) {};
        \node[circle,fill,inner sep=2pt,green] at (axis cs:11.3150, 0) {};
        
        \node[circle,fill,inner sep=2pt,blue] at (axis cs:0, 6.3040) {};
        \node[circle,fill,inner sep=2pt,red] at (axis cs:0, -6.8130) {};
        \node[circle,fill,inner sep=2pt,green] at (axis cs:0, -2.5230) {};
        \node[circle,fill,inner sep=1pt] at (axis cs:0,0) {};

        \end{axis}
    \end{tikzpicture}
    \end{lrbox}
    \resizebox{0.95\columnwidth}{!}{\usebox{\mybox}}
    \caption{\textit{Left}: Reduction in the feasible regions in the $R_c-d_c$ space (marked in red) for the example considered in Sec.~\ref{Sec:RcdcDes}; \textit{Right}: Initial positions and paths traced by the 12 agents for the solution $R_c=2400, d_c=1525$.}
    \label{fig:AddAgents}
\end{figure}
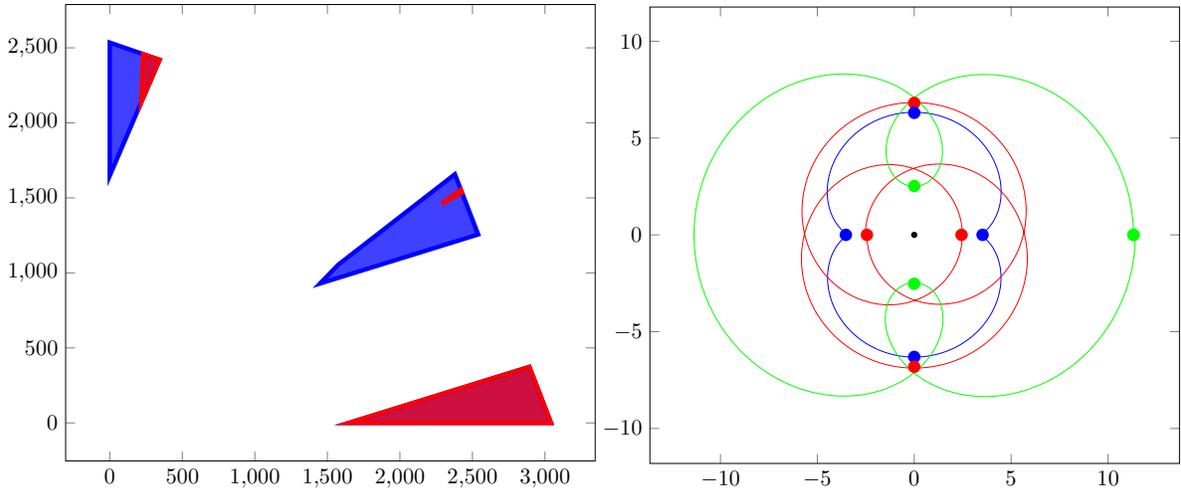

For the example considered in Sec.~\ref{Sec:RcdcDes}, the variation in the feasible regions in the $R_c-d_c$ space with the inclusion of 3 additional agents per path, and the paths traced by them, are shown in Fig.~\ref{fig:AddAgents}. As can be seen, the region in the center of the $R_c-d_c$ space has significantly shrunk, while that close to the $R_c-$ axis has not. This implies that for epitrochoidal paths that are close to circles, as shown in Fig.~\ref{fig:DiffRcDcEx}, additional agents can be injected quite easily - this is a natural result as none of these paths intersects each other, while those found from the center of the $R_c-d_c$ space self-intersect and with other paths as well. It is remarked that it may be possible to inject more than 3 agents per path, by solving the minimisation problem numerically for feasible $\phi$. As should be evident, the frequency with which a point in the 2-D space is visited increases linearly as the number of agents, thus increasing the refresh rate.

\section{Results}\label{Sec:Res}

Prior to presenting experimental results, challenges that can arise in implementation of the proposed design are identified and addressed analytically. Mainly, the issues of implementing the designed CP on non-holonomic robots with input bounds and perturbations in the designed initial positions are considered.

\subsection{Inclusion of Speed and Turn-rate Constraints}\label{Sec:NHRobSense}

As introduced in Sec.~\ref{Sec:RobCPProp}, the non-holonomic robots whose kinematics are defined in \eqref{AgentKin} have bounded linear speed and turn-rate, defined by the constraints $|V|\leq V_{\max}$ and $|\omega|\leq \omega_{\max}$, respectively. In the robot platform used for experimental evaluation, it is $V,\omega$ that should be provided as inputs; thus, considering their limits is crucial. In this section, the influence of such constraints on feasible regions in the $R_c- d_c$ space is discussed. 

For analysis, the parametric form of the epitrochoid, \eqref{xEpiAj} and \eqref{yEpiAj}, are used. In terms of the parameter $\theta$, the linear speed and turn-rate are given by
\begin{subequations}\label{AgentSpeeds}
    \begin{align}
        V(\theta) &= \pm\left(\left(\frac{dx}{d\theta}\right)^2 + \left(\frac{dy}{d\theta}\right)^2\right)^{0.5}= \pm (k+1)\left(r^2+d^2-2rd\cos{\left(k\theta\right)}\right)^{0.5}, \label{LinSpeed}
        \intertext{and}
        \omega(\theta) &= \frac{ \frac{d^2y}{d\theta^2}\frac{dx}{d\theta} -  \frac{d^2x}{d\theta^2}\frac{dy}{d\theta}}{V(\theta)^2} = \frac{r^2+d^2(k+1)-rd(k+2)\cos{\left(k\theta\right)}}{r^2+d^2-2rd\cos{\left(k\theta\right)}}. \label{AngSpeed}
    \end{align}
\end{subequations}
If the coefficients satisfy $d=r$, which is a cusp of the epitrochoid, then, at $\theta=\frac{2m\pi}{k}, \ m=0,1,\cdots$, the linear speed $V(\theta=\frac{2m\pi}{k})=0$ and the turn-rate $\omega(\theta=\frac{2m\pi}{k})$ becomes undefined. This is because, at the cusp, the robot would have to come to a stop, change its orientation, and trace the path from this new orientation. To avoid this discontinuity, if still achievable behaviour, certain values of $(R_c,d_c)$ are eliminated. These values correspond to when the coefficients $c_{ir}$ and $c_{id}$, defined in \eqref{XYlmaxminExp} and \eqref{ceqs}, become $c_{ir}=c_{id} \ \forall \ i $. As an added safety margin, these infeasible regions are expanded using the inequality 
\begin{equation}\label{CuspRangeRel}
    \left(1-\epsilon\right)\leq\Big|\frac{c_{id}}{c_{ir}}\Big|\leq\left(1+\epsilon\right), \ 0<\epsilon<1.    
\end{equation}
Thus, the feasible regions may further shrink when this constraint is added. 

At $\theta=\frac{2(m+1)\pi}{k}, \ m=0,1,\cdots$, the linear speed and turn-rate achieve their maximum, given by
\begin{subequations}\label{AgentSpeedsMax}
    \begin{align}
        V_{\theta \max} &= \pm (k+1)r\left(\frac{d}{r}+1\right), \label{LinSpeedMax} \\
        \omega_{\theta \max} &= \frac{\left(\frac{d}{r}+1\right)^2+k\frac{d}{r}\left(\frac{d}{r}+1\right)}{\left(\frac{d}{r}+1\right)^2}, \label{AngSpeedMax}
    \end{align}
\end{subequations}
respectively. Clearly, if $\frac{d}{r}\gg1$, the linear speed requirement becomes large, possibly violating the bound $V\leq V_{\max}$, although the angular speed demand may be low; the condition $\frac{d}{r}\gg1$ approximates the trajectory to a circle with radius $d$ - a large number. These solutions of $\left(R_c,d_c\right)$ may also need to be rejected. To summarise, during implementation on non-holonomic robots with bounds on their inputs, the feasible regions in the $R_c-d_c$ shrink further, as those solutions are eliminated that: \textit{i}. lead to cusps even for one of the agents; and \textit{ii}. result in high demands on the linear or angular speeds.

In the example presented in Sec.~\ref{Sec:RcdcDes}, the values of $R_c,d_c$ that lead to cusps for each agent are shown by the straight lines (by setting $\epsilon=0$ in \eqref{CuspRangeRel}) in the left sub-figure in Fig.~\ref{fig:DiffRcDcEx}. As can be seen, for Agent $A_2$ (red line), there are parts of the feasible region that lead to a discontinuity in turn-rate. The almost circular paths in the center sub-figure demand constant turn-rate and hence zero angular acceleration, while for those in the right sub-figure, larger values of turn-rate and angular acceleration are expected at the troughs of the trochoidal path. Thus, a trade-off needs to be performed in the selection of these paths. A decision variable that can be employed is the area traced by the robots moving on these paths, when they are performing a surveillance or guarding task - a larger surface area coverage can be expected with trochoidal paths that contain troughs/crests than paths that resemble circles.

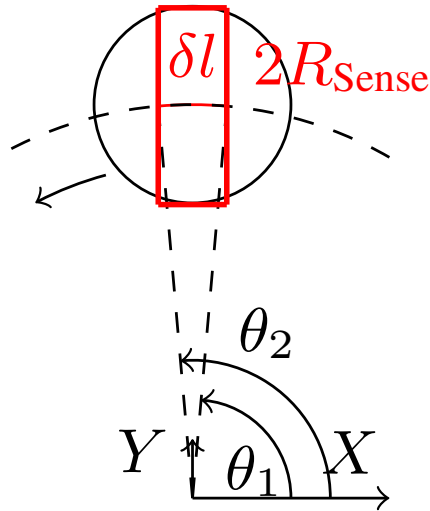
\begin{figure}[htpb!]
    \centering
    \resizebox{0.4\columnwidth}{!}{    
        \begin{tikzpicture}
            \draw[->] (0,0) -- (0,0.3) node[font=\footnotesize,pos=0.8,left] {$Y$};
            \draw[->] (0,0) -- (1,0) node[font=\footnotesize,pos=0.8,above] {$X$};            
            \draw[dashed] (0,0) -- ++(85:2cm)
                  (0,0) -- ++(95:2cm);    
            \draw[red] ([shift=(85:2cm)]0,0) arc (85:95:2cm);
            \draw[dashed] ([shift=(60:2cm)]0,0) arc (60:120:2cm);
            \draw[->] ([shift=(105:1.7cm)]0,0) arc (105:118:1.7cm);
            \draw[->] ([shift=(0:0.5cm)]0,0) arc (0:85:0.5cm) node[font=\footnotesize,pos=0.6,below] {$\theta_1$};
            \draw[->] ([shift=(0:0.7cm)]0,0) arc (0:95:0.7cm) node[font=\footnotesize,pos=0.6,above] {$\theta_2$};

             \draw (0,2) circle (0.5 cm);
             \draw[thick,red] (0.174,2.492) -- (0.174,1.492) node[font=\footnotesize,pos=0.3,right] {$2R_{\text{Sense}}$};
             \draw[thick,red] (-0.174,2.492) -- (-0.174,1.492);
             \draw[thick,red] (-0.174,2.492) -- (0.174,2.492) node[font=\footnotesize,pos=0.5,below] {$\delta l$};
             \draw[thick,red] (-0.174,1.492) -- (0.174,1.492);

        \end{tikzpicture}
    }
    \caption{Computing the area covered by a robot with sensing radius $R_{\text{Sense}}$ tracing a path}
    \label{fig:AreaCov}
\end{figure}

\textit{Area Coverage}: To design the paths, in Sec.~\ref{Sec:RcdcDes}, an agent is modeled as being bounded in size by a circle of radius $R_{\text{Rob}}$. Further, assume that each agent is also equipped with a range sensor with range $R_{\text{Sense}}\gg R_{\text{Rob}}$ and 360$^\circ$ sensing around the robot (for instance, with a LIDAR sensor or equivalent). Now, from the schematic shown in Fig.~\ref{fig:AreaCov}, the sensing area covered by an agent $A_i$ in one period of the trochoidal path can be calculated. The infinitesimally small area (in red) is given by the product $2R_{\text{Sense}}\delta l_i$, where $\delta l_i=\sqrt{\left(\dfrac{\partial x_i}{\partial \theta}\right)^2 + \left(\dfrac{\partial y_i}{\partial \theta}\right)^2}$ is the length of the trochoidal path (expressed in the parametric form) traced for an infinitesimally small angle $\delta \theta=\left(\theta_2-\theta_1\right)$. Thus, as the trochoids are periodic and closed, the total sensing area covered by agent $A_i$ is given by 
\begin{equation}\label{AreaAgentAi}
    \text{SenseArea}_i = 2R_{\text{Sense}}\int\limits_{0}^{2\pi} \sqrt{\left(\frac{\partial x_i}{\partial \theta}\right)^2 + \left(\frac{\partial y_i}{\partial \theta}\right)^2}d\theta.
\end{equation}
Thus, the total area covered by all 3 agents is simply the sum $\sum\limits_{i=1}^{3}{\text{SenseArea}_i}$. 

From the feasible regions shown in Fig.~\ref{fig:DiffRcDcEx}, for the design pair $\left(R_c, d_c \right) = (2500,0)$, which leads to circular paths, the total sensing area is found to be 146.98 (determined numerically\footnote{\url{https://in.mathworks.com/help/matlab/ref/integral.html}}), whereas for $\left(R_c, d_c \right) = (2000,1200)$, which results in epitrochoids with pronounced troughs/crests, the total area is 243.84 (all in appropriate units). Thus, a considerable increase in sensing area can be achieved with the use of trochoidal paths, which may require operating close to the inputs' limits of the individual agents, than circles, which do not demand such input magnitudes.

\subsection{Perturbing Initial Positions}\label{Sec:PertIC}

In Sec.~\ref{Sec:InitPos}, it is shown that feasible epitrochoidal paths can be generated by choosing the agents' initial positions to lie on the $X-$axis, say $(x_{i0F},0)$. The effect of perturbations in these initial positions is now analysed; the other design parameters remain unchanged. Let the perturbed position of each agent be given by $(x_{i0F}+\epsilon_i\cos\psi_i,\epsilon_i\sin\psi_i)$, where, $\epsilon_i>0, \ \psi_i\in[0,2\pi] \ \forall \ i$. These new initial positions lead to different values of $(R_c,d_c)$, denoted by $(R_{cp},d_{cp})$, hence, different trochoidal paths, and also for the paths to evolve about a new CoR, since, from \eqref{rcdcEqs}, $\phi_{r,d,0}\neq0$, and hence, in \eqref{XYlmaxminExp}, $c_0=c_{i0}=c_{j0}\neq0$. 

For the example considered in Sec.~\ref{Sec:RcdcDes}, the initial positions that are found for the solution $R_c=2500,d_c=0$ (black circle in the left sub-figure in Fig.~\ref{fig:DiffRcDcEx}) are perturbed (with $\epsilon_i=1 \ \forall \ i$ and $\psi_i$ randomly chosen to lie in $[0,2\pi]$). These perturbed initial positions result in the $R_c,d_c$ pair marked by the green circle in the same region, which in turn lead to different trochoidal paths (dashed lines in the center sub-figure in Fig.~\ref{fig:DiffRcDcEx}); with perturbations, the path closest to the CoR exhibits a clear epitrochoidal feature, which is different from the circular path generated in the absence of any perturbations. From this illustration, the importance of finding ``large'' regions in the $R_c-d_c$ space can be appreciated, as such ``large'' regions will hold solutions for feasible paths even in the presence of perturbed initial positions.

It can be shown that since $(R_{cp},d_{cp})$ and the new CoR is common to all agents, the smallest and largest distances between any pair of agents continue to satisfy the inequalities \eqref{PairCloseDistIneq} and \eqref{PairFarDistIneq}, respectively. Thus, perturbing the initial positions of the agents does not lead to collisions between them. On the other hand, the smallest distance-to-the-origin inequality, \eqref{CloseDistIneq}, may be violated. Bounds on the magnitudes of perturbation, $\epsilon_i$, are now derived so that \eqref{CloseDistIneq} still holds. 

This analysis is performed for a closed epitrochoid with the origin as the CoR and expressed in the parametric form $x_\theta = \left(k+1\right)rc_{\theta} - dc_{(k+1)\theta}, \ y_\theta = \left(k+1\right)rs_{\theta} - ds_{(k+1)\theta}$, where $c_{(\cdot)}=\cos{(\cdot)},s_{(\cdot)}=\sin{(\cdot)}$. A point on this epitrochoid is closest to the origin at $\theta=0,\pi$ with distance $(r(k+1)-d)$. Now, let the initial coordinates, evaluated at $\theta=0$, be perturbed such that the CoR of the perturbed epitrochoid (PE) has the coordinates $(\delta\cos{\gamma_T},\delta\sin{\gamma_T}), \ \delta>0$ and its orientation is changed by an angle $\gamma_R$. Thus, the parametric form of the PE becomes
\begin{align*}
    x_{\theta P} &= \left(k+1\right)rc_{\theta+\gamma_R} - dc_{(k+1)\theta+\gamma_R} + \delta c_{\gamma_T},  \nonumber \\
    y_{\theta P} &= \left(k+1\right)rs_{\theta+\gamma_R} - ds_{(k+1)\theta+\gamma_R} + \delta s_{\gamma_T}.
\end{align*}
The squared distance of any point on the PE to the origin is given by
\begin{align*}
    d_{eP}^2 &= \left(k+1\right)^2r^2 + d^2  - 2(k+1)rd c_{k\theta} \\
             & + \delta^2 + 2\delta\left((k+1)rc_{\left(\gamma_R-\gamma_T+\theta\right)} - dc_{\left(\gamma_R-\gamma_T+(k+1)\theta\right)} \right).\nonumber 
\end{align*}
As the magnitudes of the perturbation is expected to be small, its influence on $d_{eP}^2$ is analysed by linearising this expression about the following points: \textit{i}. $\delta=\gamma_{R,T}=\theta=0$ and \textit{ii}. $\delta=\gamma_{R,T}=0$ and $\theta=\pi$. It can be seen that these are the points when the original epitrochoid is closest to the origin. It can be shown that in the vicinity of $\theta=0$, $d_{eP}^2\leq\left(r(k+1)-d+\delta\right)^2$, while in the vicinity of $\theta=\pi$, $d_{eP}^2\leq\left(r(k+1)-d-\delta\right)^2$. Thus, the PE shifts closer to the origin at $\theta=\pi$ and farther away at $\theta=0$; in both cases, by no more than $\delta$. This shift in the paths can be observed in the dashed lines in the center sub-figure in Fig.~\ref{fig:DiffRcDcEx}.

Now, the point at $\theta=0$ on the PE has the coordinates $\left[\left(\left(r(k+1)-d\right)c_{\gamma_R}+\delta c_{\gamma_T}\right),\left(\left(r(k+1)-d\right)s_{\gamma_R}+\delta s_{\gamma_T}\right)\right]$. Therefore, for small $\gamma_{R,T}$, these coordinates can be approximated to $\left[\left(r(k+1)-d+\delta\right),\left(r(k+1)-d\right)\gamma_R\right]$. Thus, the magnitude of perturbation $\epsilon_i\approx\left(\delta^2+\left(r(k+1)-d\right)^2\gamma_R^2\right)^{0.5}$; the angle $\psi_i\approx\frac{\left(r(k+1)-d\right)\gamma_R}{\delta}$. The implication of perturbed initial conditions is that the agents can enter the guarded region around the asset, thus violating one of the key design requirements. To avoid this issue, the inequality \eqref{CloseDistIneq} is amended to
\begin{equation}\label{CloseDistIneqMod}
    d_{i0\min} = \big||\alpha_{ir}|R_c  -  |\alpha_{id}|d_c\big|\geq \left(d_{0\min}+\delta\right),
\end{equation}
so that perturbations of magnitude $\epsilon_i$ in the initial positions do not lead to any constraint violations. While the farthest distance inequality \eqref{FarDistIneq} should be similarly modified, if $d_{i0\max}\gg\delta$, then, these perturbations do not affect this constraint greatly.

\subsection{Hardware Implementation}

The proposed algorithms are implemented on the QBOT 2E mobile robot platform by Quanser\footnote{\url{https://www.quanser.com/products/qbot-2e/}}. All experiments are conducted in the Autonomous Vehicles Research Studio\footnote{\url{https://www.quanser.com/products/autonomous-vehicles-research-studio/}}, also from Quanser, using 3 robots. The Studio allows for inter-robot communication, with which robots exchange their coordinates with their neighbours and hence the CP, \eqref{CPExp}, can be implemented; the communication topology as shown in Fig.~\ref{fig:3AgentFig} is adopted. The robots' coordinates, including orientation, $\gamma_i$, are measured by a tracking system built-in to the Studio. To control the motion of the robots, their linear and angular velocities need to be provided. The angular velocity is implemented as the proportional-integral controller with feed-forward term, given by
\begin{equation*}
    \omega_{i}=K_P\left(\gamma_{i\text{Ref}}-\gamma_{i}\right) + K_I\int_{0}^{t}\left(\gamma_{i\text{Ref}}-\gamma_{i}\right)dt + \left(\frac{d\gamma_{i\text{Ref}}}{dt}\right),
\end{equation*}
where $K_P>0$ and $K_I>0$ are the gains; the linear speeds, reference orientation angles and their time-derivatives are determined using the formulae presented in Sec.~\ref{Sec:RobCPProp}.

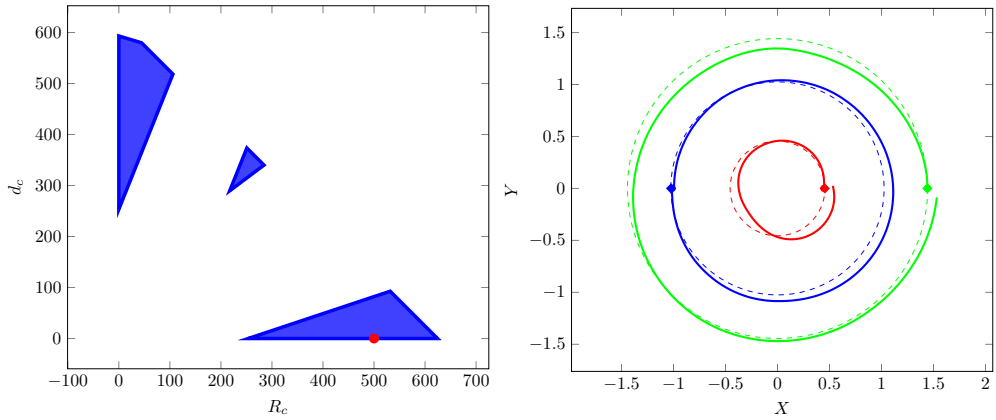
\begin{figure}[!tbp]
  \centering
    \begin{lrbox}{\mybox}%

    \begin{tikzpicture}
        \begin{axis}[scale only axis, axis equal, xlabel= $R_c$, ylabel = $d_c$]
            \addplot[name path = t1, line width=2pt, color = blue] table {
            217.2857    289.7143
            250.8316    373.5789
            284.3774    340.0331
            } -- cycle;
            \addplot[name path = t2, line width=2pt, color = blue] table {
            44.6008     579.8098
            105.9744    518.4361
            0   253.5000
            0   593.1900
            } -- cycle;
            \addplot[name path = t3, line width=2pt, color = blue] table {
            531.6829   92.7276
            624.4105    0
            253.5000    0
            } -- cycle;
            \addplot[thick, color = blue, fill = blue, fill opacity = 0.5]fill between[of=t1 and t2];
            \addplot[thick, color = blue, fill = blue, fill opacity = 0.5]fill between[of=t2 and t3];
            \addplot[thick, color = blue, fill = blue, fill opacity = 0.5]fill between[of=t3 and t1];
            \node[circle,fill,inner sep=2pt,red] at (axis cs:500, 0) {};

        \end{axis}
    \end{tikzpicture}
    
    \begin{tikzpicture}
    \begin{axis}[axis equal, scale only axis, xlabel= $X$, ylabel = $Y$]
        \addplot[name path = A1, color = blue, dashed] table{A1_CIRC_ED_2.txt};
        \addplot[name path = A2, color = red, dashed] table{A2_CIRC_ED_2.txt};
        \addplot[name path = A3, color = green, dashed] table{A3_CIRC_ED_2.txt};
        
        \addplot[name path = AR1, color = blue, very thick] table{A1_RCIRC_ED.txt};
        \addplot[name path = AR2, color = red, very thick] table{A2_RCIRC_ED.txt};
        \addplot[name path = AR3, color = green, very thick] table{A3_RCIRC_ED.txt};

        \node[label={},diamond,fill,blue,inner sep=1.5pt] at (axis cs:-1.0240,0){};
        \node[label={},circle,fill,blue,inner sep=1.5pt] at (axis cs:-1.0142,0.001){};

        \node[label={},diamond,fill,red,inner sep=1.5pt] at (axis cs:0.4550,0){};
        \node[label={},circle,fill,red,inner sep=1.5pt] at (axis cs:0.4473,0.0009){};

        \node[label={},diamond,fill,green,inner sep=1.5pt] at (axis cs:1.4410,0){};
        \node[label={},circle,fill,green,inner sep=1.5pt] at (axis cs:1.4422,0.0009){};
        \end{axis}
    \end{tikzpicture}%
    
    \end{lrbox}
    \resizebox{0.8\columnwidth}{!}{\usebox{\mybox}}
    \caption{Design parameters: $k = 2$, $s_{1,2,3}=\left(5,12,13\right)$, $d_{0\min}$ = 0.01, $d_{0\max}$ = 1.8, $d_{CR}= 4$, and $d_{CT} = 0.5$. \textit{Left}: Feasible regions in the $R_c-d_c$ space; \textit{Right}: Reference (dashed) and Actual (solid) Paths for $R_c=500,d_c=0$. Blue is Agent $A_1$, red is Agent $A_2$, and green is Agent $A_3$.}
    \label{fig:Exp_Circ}
\end{figure}

\begin{figure}[!tbp]
  \centering
    \begin{lrbox}{\mybox}%

    \begin{tikzpicture}
        \begin{axis}[scale only axis, axis equal, xlabel= $R_c$, ylabel = $d_c$]
            \addplot[name path = t1, line width=2pt, color = blue] table {
            932.2967    1624.5
            1271.3  1285.5
            987.9   1002.1
            661.8   808.8
            912.8   1589.9
            } -- cycle;
            \addplot[name path = t2, line width=2pt, color = blue] table {
            1285.5  1271.3
            1343.8  1213
            1057.6  1043.4
            } -- cycle;
            \addplot[name path = t3, line width=2pt, color = blue] table {
            229.9   1965.1
            0   1250
            0   2008.9
            } -- cycle;
            \addplot[name path = t4, line width=2pt, color = blue] table {
            2157.1  399.7
            2556.8  0
            937.5   0
            2109.4  390.6
            } -- cycle;
            \addplot[thick, color = blue, fill = blue, fill opacity = 0.75]fill between[of=t1 and t2];
            \addplot[thick, color = blue, fill = blue, fill opacity = 0.75]fill between[of=t3 and t4];
            \node[circle,fill,inner sep=2pt,red] at (axis cs:1000, 1250) {};
        \end{axis}
    \end{tikzpicture}
    
    \begin{tikzpicture}
    \begin{axis}[axis equal, scale only axis, xlabel= $X$, ylabel = $Y$]
        \addplot[name path = A1, color = blue, dashed] table{A1_TROCH_ED_2.txt};
        \addplot[name path = A2, color = red, dashed] table{A2_TROCH_ED_2.txt};
        \addplot[name path = A3, color = green, dashed] table{A3_TROCH_ED_2.txt};
        
        \addplot[name path = AR1, color = blue, very thick] table{A1_RTROCH_ED.txt};
        \addplot[name path = AR2, color = red, very thick] table{A2_RTROCH_ED.txt};
        \addplot[name path = AR3, color = green, very thick] table{A3_RTROCH_ED.txt};

        \node[label={},circle,fill,blue,inner sep=1.5pt] at (axis cs:-0.16,0){};
        \node[label={},star,fill,blue,inner sep=1.5pt] at (axis cs:-0.1610,0.0042){};

        \node[label={},circle,fill,red,inner sep=1.5pt] at (axis cs:-0.9490,0){};
        \node[label={},star,fill,red,inner sep=1.5pt] at (axis cs:-0.9400,0.0007){};

        \node[label={},circle,fill,green,inner sep=1.5pt] at (axis cs:1.5840,0){};
        \node[label={},star,fill,green,inner sep=1.5pt] at (axis cs:1.5884,0.0011){};
        
        \end{axis}
    \end{tikzpicture}%
    
    \end{lrbox}
    \resizebox{0.8\columnwidth}{!}{\usebox{\mybox}}
    \caption{Design parameters: $k = 2$, $s_{1,2,3}=\left(7,24,25\right)$, $d_{0\min}$ = 0.01, $d_{0\max}$ = 1.8, $d_{CR}= 4$, and $d_{CT} = 0.5$. \textit{Left}: Feasible regions in the $R_c-d_c$ space; \textit{Right}: Reference (dashed) and Actual (solid) Paths for $R_c=1000,d_c=1250$. Blue is Agent $A_1$, red is Agent $A_2$, and green is Agent $A_3$.}
    \label{fig:Exp_Troch}
\end{figure}
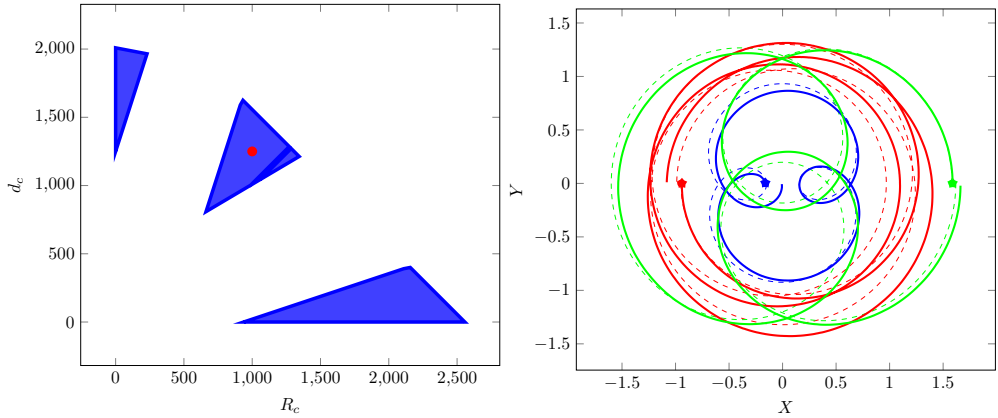

\begin{figure}[!tbp]
  \centering
    \begin{lrbox}{\mybox}%

        \begin{tikzpicture}
            \begin{axis}[scale only axis, xlabel= Time (sec), ylabel = Distances (m),legend style={draw=none},xmin=0.0,xmax=350,ymin=0,ymax=2.5]
                \addplot[name path = A1, color = blue, line width = 2 pt] table [x index=0,y index=1] {IAOD.txt};
                \addlegendentry{Agent $A_1$}
                \addplot[name path = A2, color = red, line width = 2 pt] table [x index=0,y index=2] {IAOD.txt};
                \addlegendentry{Agent $A_2$}
                \addplot[name path = A3, color = green, line width = 2 pt] table [x index=0,y index=3] {IAOD.txt};
                \addlegendentry{Agent $A_3$}
                 \addplot [black, dashed, line width = 2 pt,domain=0:350,samples=200]{0.01};
                \addlegendentry{$d_{0\min}=0.01$}
                \addplot [black, dashed, line width = 2 pt,domain=0:350,samples=200]{1.8};
				\addlegendentry{$d_{0\max}=1.8$}

            \end{axis}
        \end{tikzpicture}%
    
        \begin{tikzpicture}
            \begin{axis}[scale only axis, xlabel= Time (sec), ylabel = Distances (m),legend style={draw=none},xmin=0.0,xmax=350,ymin=0,ymax=4.5]
                \addplot[name path = A1, color = cyan, line width = 2 pt] table [x index=0,y index=4] {IAOD.txt};
                \addlegendentry{$A_1--A_2$}
                \addplot[name path = A2, color = magenta, line width = 2 pt] table [x index=0,y index=5] {IAOD.txt};
                \addlegendentry{$A_1--A_3$}
                \addplot[name path = A3, color = lime, line width = 2 pt] table [x index=0,y index=6] {IAOD.txt};
                \addlegendentry{$A_2--A_3$}
                \addplot [black, dashed, line width = 2 pt,domain=0:350,samples=200]{0.5};
				\addlegendentry{$d_{CT}=0.5$}
                \addplot [black, dashed, line width = 2 pt,domain=0:350,samples=200]{4};
				\addlegendentry{$d_{CR}=4$}
            
            \end{axis}
        \end{tikzpicture}%
    
    \end{lrbox}
    \resizebox{0.8\columnwidth}{!}{\usebox{\mybox}}
    \caption{\textit{Left}: Temporal variation in the distances of the robots from the origin; \textit{Right}: Temporal variation of the pair-wise distances between the robots. The path parameters are based on the values shown in Fig.~\ref{fig:Exp_Troch}.}
    \label{fig:TDom_Troch}
\end{figure}
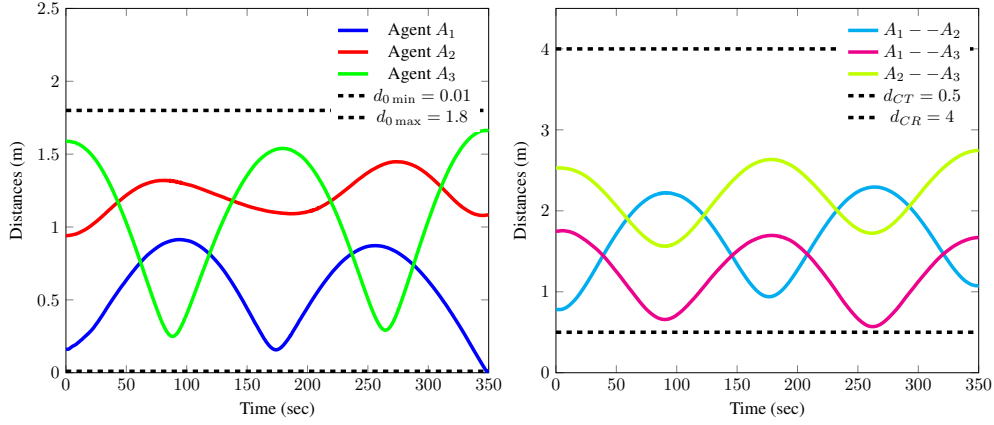

The robots move in an environment which is a square of size $2\times2 \ \text{m}^2$; the robots themselves are circles with radius, $R_{\text{Rob}}=0.015$ m. The distances used in the inequalities \eqref{IneqExp} are set as $d_{0\min} = 0.01$, $d_{0\max} = 1.8$, $d_{CR}= 4$, and $d_{CT} = 0.5$ (all in m). Owing to the relatively large size of the robots with respect to the environment and to obtain ``reasonable'' paths, it became necessary to choose $d_{0\min} = 0.01$. Similar to the results shown in Fig.~\ref{fig:DiffRcDcEx}, two classes of paths are generated which the robots trace: \textit{i}. circular and \textit{ii}. epitrochoidal. The reference (dashed lines) and actual paths (solid lines) in the $X-Y$ plane for these two cases are shown in Figs.~\ref{fig:Exp_Circ} and \ref{fig:Exp_Troch}, respectively. The evolution in time of the distances of the three agents from the origin as well as inter-robot pair-wise distances are shown for the epitrochoidal path in Fig.~\ref{fig:TDom_Troch}. As can be seen, all distance constraints listed in \eqref{IneqExp} are satisfied by the agents. Thus, no pair of agents collide with each other and neither do they trace large paths that would cause them to cross the boundaries of the environment.	

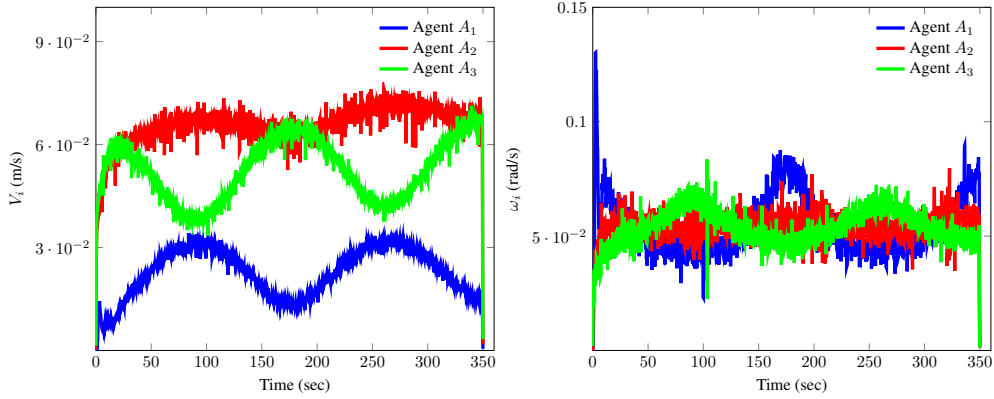
\begin{figure}[!tbp]
  \centering
    \begin{lrbox}{\mybox}%

        \begin{tikzpicture}
            \begin{axis}[scale only axis, xlabel= Time (sec), ylabel = $V_i$ (m/s), ytick={0.03,0.06,0.09},legend style={draw=none},xmin=0.0,xmax=360,ymin=0,ymax=0.1] 
                \addplot[name path = AV1, color = blue, line width = 2 pt] table [x index=0,y index=1] {T_TVW_A1_sdC.txt};
                \addlegendentry{Agent $A_1$}
                \addplot[name path = A2, color = red, line width = 2 pt] table [x index=0,y index=1] {T_TVW_A2_sdC.txt};
                \addlegendentry{Agent $A_2$}
                \addplot[name path = A3, color = green, line width = 2 pt] table [x index=0,y index=1] {T_TVW_A3_sdC.txt};
                \addlegendentry{Agent $A_3$}

            \end{axis}
        \end{tikzpicture}%
    
        \begin{tikzpicture}
            \begin{axis}[scale only axis, xlabel= Time (sec), ylabel = $\omega_i$ (rad/s), ytick={0.05,0.1,0.15}, legend style={draw=none},xmin=0.0,xmax=360,ymin=0,ymax=0.15] 
                \addplot[name path = AV1, color = blue, line width = 2 pt] table [x index=0,y index=2] {T_TVW_A1_sdC.txt};
                \addlegendentry{Agent $A_1$}
                \addplot[name path = A2, color = red, line width = 2 pt] table [x index=0,y index=2] {T_TVW_A2_sdC.txt};
                \addlegendentry{Agent $A_2$}
                \addplot[name path = A3, color = green, line width = 2 pt] table [x index=0,y index=2] {T_TVW_A3_sdC.txt};
                \addlegendentry{Agent $A_3$}

            \end{axis}
        \end{tikzpicture}%
    
    \end{lrbox}
    \resizebox{0.8\columnwidth}{!}{\usebox{\mybox}}
    \caption{\textit{Left}: Temporal variation in the linear speeds of the robots; \textit{Right}: Temporal variation of the turn-rates the robots. The path parameters are based on the values shown in Fig.~\ref{fig:Exp_Troch}.}
    \label{fig:TDom_TrochSpeed}
\end{figure}

The design variables are presented in Table~\ref{tab:PathParamVal}. As mentioned in the Introduction, the proposed CP design procedure simplifies the selection of 10 variables to just 4. To generate these results, the robots were driven to their respective initial positions prior to the implementation of the designed CP. The reference and actual initial positions of the three agents are also listed in Table~\ref{tab:PathParamVal}; as these are slightly different, the paths traced by the robots also differ from the designed ones, but are still feasible. Further, the magnitudes of the Pythagorean triples, denoted by $s_{1,2,3}$, in Table~\ref{tab:PathParamVal} were scaled \textit{down} by a factor of 0.01 for the circular path and by a factor of 0.0015 for the epitrochoidal path; with this scaling down of the speed, the robots took close to 6 min to trace one period of the path. As the robot platform allows for only the linear speed and turn-rate as the inputs to each robot, instead of controlling the input to each wheel of the robot, this scaling down had to be performed to ensure the robots trace the designed paths; the time-histories of the linear speeds and turn-rates are presented in Fig.~\ref{fig:TDom_TrochSpeed}. Note that the analysis in Sec.~\ref{Sec:NHRobSense} focussed on ensuring that paths with cusps are avoided; the upper-limits on speed and turn-rate constraints are dependent on the robot platform.

\begin{table}[htpb!]  
	\centering
	\caption{Parameters used to generate the paths shown in Figs.~\ref{fig:Exp_Circ} and \ref{fig:Exp_Troch}.}
    \label{tab:PathParamVal}
      \begin{tabular}{p{0.17\columnwidth}|p{0.25\columnwidth}|p{0.44\columnwidth}}
    		\toprule
    		Path & Design Variables & Reference and Actual (in []) initial positions \\ \midrule
    		\multirow{3}*{\parbox{0.2\columnwidth}{Circular
                          (Fig.~\ref{fig:Exp_Circ})}} & \multirow{3}*{$s_{1,2,3}=\left(5,12,13\right)$} & $(-1.024,0) \ [-1.014;0.001]$ \\
                                                              &                                                          & $(0.455;0) \ [0.447;0.001]$ \\
                                                              &                                                          & $(1.441;0) \ [1.442;0.001]$ \\ \midrule
            \multirow{3}*{\parbox{0.2\columnwidth}{Epitrochoidal
                          (Fig.~\ref{fig:Exp_Troch})}} & \multirow{3}*{$s_{1,2,3}=\left(7,24,25\right)$} & $(-0.16,0) \ [-0.161;0.004]$ \\
                                                              &                                                          & $(-0.949;0) \ [-0.94;0.001]$ \\
                                                              &                                                          & $(1.584;0) \ [1.588;0.001]$ \\ \midrule          
            \multicolumn{3}{c}{$k=2$, $d_{0\min}$ = 0.01, $d_{0\max}$ = 1.8, $d_{CR}= 4$, and $d_{CT} = 0.5$} \\ 
    		\bottomrule     
      \end{tabular}  
\end{table}

As can be seen in the left sub-figure in Fig.~\ref{fig:Exp_Circ}, for the chosen Pythagorean triple of $s_{1,2,3}=\left(5,12,13\right)$, the region in the $R_c-d_c$ space that would yield a non-circular path is much smaller than the other two regions; as discussed in Sec.~\ref{Sec:PertIC}, selecting a solution for $(R_c,d_c)$ from this ``small region'' would not allow for feasible paths to be traced if the initial positions are perturbed. On the other hand, from the left sub-figure in Fig.~\ref{fig:Exp_Troch}, as the corresponding region is larger, greater perturbations in the initial positions can be tolerated. Thus, for each class of path, the $(R_c,d_c)$ pair is selected (red circles in both sub-figures), so that perturbations can be handled.

\begin{figure}[htpb!]
    \centering
    \begin{lrbox}{\mybox}%
    \begin{tikzpicture}
                \begin{axis}[axis equal, scale only axis, xlabel= $R_c$, ylabel = $d_c$, legend style={draw=none, legend columns = 1, nodes={scale=0.9}} ]
                \addplot[name path = t1_2, color = blue, fill = blue, area legend] coordinates{(0.1129, 0.7957) (0, 0.5417) (0, 0.8333)} --cycle;
                \addlegendentry{$\mathcal{S}_1$}
                \addplot[name path = t2_3, color = red, fill = red, area legend] coordinates{(0.0649, 0.8672) (0, 0.6869) (0, 0.8889)} --cycle; 
                \addlegendentry{$\mathcal{S}_2$}
                \addplot[name path = t3_3, color = green, fill = green, area legend] coordinates{(0.1481, 0.7506) (0, 0.4604) (0, 0.8000)} --cycle;
                \addlegendentry{$\mathcal{S}_3$}
                \addplot[name path = i12_23, color = magenta, fill = magenta, area legend] table{
                     0    0.8333
                0.0471    0.8176
                     0    0.6869
                }--cycle;
                \addlegendentry{$\mathcal{S}_1\cap\mathcal{S}_2$}
                \addplot[name path = i12_33, color = cyan, fill = cyan, area legend] table{
                     0    0.5417
                     0    0.8000
                0.1000    0.7666
                }--cycle;
                \addlegendentry{$\mathcal{S}_1\cap\mathcal{S}_3$}
                \addplot[name path = i23_33, color = yellow, fill = yellow, area legend] table{
                     0    0.6869
                     0    0.8000
                0.0363    0.7879
                }--cycle;
                \addlegendentry{$\mathcal{S}_2\cap\mathcal{S}_3$}
                \addplot[name path = i12_23_33, color = black, fill = black, area legend] table{
                     0    0.6869
                     0    0.8000
                0.0363    0.7879
                }--cycle;
                \addlegendentry{$\mathcal{S}_1\cap\mathcal{S}_2\cap\mathcal{S}_3$}

                \addplot[name path = t1_1, color = blue, fill = blue] coordinates{(0.4765, 0.3055) (0.5175, 0.3465) (0.7821, 0.5449) (0.8348, 0.4129)} --cycle; 
                \addplot[name path = t1_3, color = blue, fill = blue] coordinates{(0.9509, 0.1228) (1.0000, 0) (0.5417, 0)} --cycle; 
            
                \addplot[name path = t2_1, color = red, fill = red] coordinates{(0.6116, 0.4437) (0.7418, 0.5739) (0.7705, 0.5099)} --cycle;
                \addplot[name path = t2_2, color = red, fill = red] coordinates{(0.4808, 0.6487) (0.5065, 0.7201) (0.5407, 0.7086)} --cycle; 
                \addplot[name path = t2_4, color = red, fill = red] coordinates{(0.9137, 0.1918) (1.0000, 0) (0.4533, 0)} --cycle; 

                \addplot[name path = t3_1, color = green, fill = green] coordinates{(0.6100, 0.4358) (0.7306, 0.5565) (0.7730, 0.5423)} --cycle; 
                \addplot[name path = t3_2, color = green, fill = green] coordinates{(0.4413, 0.2510) (0.8216, 0.4995) (0.8743, 0.3520)} --cycle; 
                \addplot[name path = t3_4, color = green, fill = green] coordinates{(0.9719, 0.0788) (1.0000, 0) (0.6343, 0)} --cycle; 
                \addplot[name path = t3_5, color = green, fill = green] coordinates{(0.9922, 0.0268) (1.0000, 0) (0.8254, 0.0000)} --cycle; 

                \addplot[name path = i11_21, color = magenta, fill = magenta] table{
                    0.7705    0.5099
                    0.6916    0.4770
                    0.7617    0.5296
                }--cycle;


                \addplot[name path = i13_24, color = magenta, fill = magenta] table{
                    0.5417         0
                    0.9455    0.1212
                    1.0000         0
                }--cycle;
                
                \addplot[name path = i11_32, color = cyan, fill = cyan] table{
                    0.8348    0.4129
                    0.5656    0.3322
                    0.8046    0.4884
                }--cycle;
                
                \addplot[name path = i13_34, color = cyan, fill = cyan] table{
                    1.0000         0
                    0.6343         0
                    0.9688    0.0781
                }--cycle;
                \addplot[name path = i13_34, color = cyan, fill = cyan] table{
                    1.0000         0
                    0.8254         0
                    0.9895    0.0264
                }--cycle;
                \addplot[name path = i21_31, color = yellow, fill = yellow] table{
                    0.7306    0.5565
                    0.7530    0.5490
                    0.7598    0.5337
                    0.6406    0.4558
                    0.6224    0.4482
                }--cycle;
                \addplot[name path = i24_34, color = yellow, fill = yellow] table{
                    1.0000         0
                    0.6343         0
                    0.9652    0.0772
                }--cycle;
                \addplot[name path = i24_35, color = yellow, fill = yellow] table{
                    1.0000         0
                    0.8254         0
                    0.9882    0.0262
                }--cycle;
                
                \addplot[name path = i13_24_34, color = black, fill = black] table{
                    1.0000         0
                    0.6343         0
                    0.9652    0.0772
                }--cycle;
                \addplot[name path = i13_24_35, color = black, fill = black] table{
                    1.0000         0
                    0.8254         0
                    0.9882    0.0262
                }--cycle;
            \end{axis}
        \end{tikzpicture}%
        \begin{tikzpicture}
                \begin{axis}[axis equal, scale only axis, xlabel= $R_c$, ylabel = $d_c$, legend style={draw=none, legend columns = 1, nodes={scale=0.9}} ]
                \addplot[name path = k2_1, color = blue, fill = blue, area legend] coordinates{(0.1129, 0.7959) (0, 0.5417) (0, 0.8333)} --cycle;
                \addlegendentry{$k_1=2$}
                \addplot[name path = k3_1, color = red, fill = red, area legend] coordinates{(0.4390, 0.2912) (0.6182, 0.4705) (0.6667, 0.4583) (0.5175, 0.3465) (0.4765, 0.3055) (0.5, 0.3125) (0.4312, 0.2738)} --cycle;
                \addlegendentry{$k_2=3$}
                \addplot[name path = k4_1, color = green, fill = green, area legend] coordinates{(0.7083, 0.2937) (0.7821, 0.3269) (0.7882, 0.3177)} --cycle;
                \addlegendentry{$k_3=4$}
                \addplot[name path = k23_1, color = magenta, fill = magenta, area legend] coordinates{(0.4765, 0.3055) (0.5175, 0.3465) (0.6667, 0.4583) (0.7308, 0.4423) (0.5, 0.3125)} --cycle;
                \addlegendentry{$k_1\cap k_2$}
                \addplot[name path = k34_1, color = yellow, fill = yellow, area legend] coordinates{(0.6020, 0.3796) (0.6148, 0.3770) (0.5963, 0.3667)} --cycle;
                \addlegendentry{$k_2\cap k_3$}
                \addplot[name path = k24_1, color = cyan, fill = cyan, area legend] coordinates{(0.6020, 0.3796) (0.6750, 0.3650) (0.5833, 0.3375)} --cycle;
                \addlegendentry{$k_1\cap k_3$}
                
                \addplot[name path = k234_1, color = black, fill = black, area legend] coordinates{(0.6020, 0.3796) (0.6148, 0.3770) (0.5963, 0.3667)} --cycle;
                \addlegendentry{$k_1\cap k_2\cap k_3$}
                \addplot[name path = k2_2, color = blue, fill = blue] coordinates{(0.9509, 0.1228) (1, 0) (0.9368, 0.1185)} --cycle;
                \addplot[name path = k2_3, color = blue, fill = blue] coordinates{(0.7821, 0.4087) (0.7381, 0.3839) (0.7874, 0.3987)} --cycle;
                \addplot[name path = k2_4, color = blue, fill = blue] coordinates{(0.7821, 0.5449) (0.8348, 0.4129) (0.5, 0.3125) (0.5963, 0.3667) (0.5833, 0.3375) (0.6750, 0.3650) (0.6148, 0.3770) (0.7308, 0.4423) (0.6667, 0.4583)} --cycle;
                \addplot[name path = k3_2, color = red, fill = red] coordinates{(0.3611,0) (0.4444, 0.0250) (0.4333, 0)} --cycle;
                \addplot[name path = k3_3, color = red, fill = red] coordinates{(0.5317, 0.2679) (0.5593, 0.2833) (0.5556, 0.2750)} --cycle;
                \addplot[name path = k3_4, color = red, fill = red] coordinates{(0.7381, 0.3839) (0.7874, 0.3987) (0.8123, 0.3520) (0.5556, 0.2750) (0.7308, 0.3538) (0.6967, 0.3607)} --cycle;
                \addplot[name path = k3_5, color = red, fill = red] coordinates{(0.9119, 0.1652) (0.9368, 0.1185) (0.9236, 0.1146) (0.8935, 0.1597)} --cycle;
                \addplot[name path = k4_2, color = green, fill = green] coordinates{(0.4583, 0.0562) (0.6458, 0.1125) (0.8922, 0.1618) (0.8935, 0.1597) (0.4444, 0.0250)} --cycle;
                \addplot[name path = k4_3, color = green, fill = green] coordinates{(0.5833, 0.3375) (0.6750, 0.3650) (0.6967, 0.3607) (0.5593, 0.2833)} --cycle;
                \addplot[name path = k23_2, color = magenta, fill = magenta] coordinates{(0.5417, 0) (0.9368, 0.1185) (1, 0)} --cycle;
                \addplot[name path = k23_3, color = magenta, fill = magenta] coordinates{(0.7821, 0.4087) (0.7874, 0.3987) (0.7381, 0.3839)} --cycle;
                \addplot[name path = k24_2, color = cyan, fill = cyan] coordinates{(0.5417, 0) (0.9236, 0.1146) (1, 0)} --cycle;
                \addplot[name path = k34_2, color = yellow, fill = yellow] coordinates{(0.7308, 0.3538) (0.5556, 0.2750) (0.5593, 0.2833) (0.6967, 0.3607)} --cycle;
                \addplot[name path = k34_3, color = yellow, fill = yellow] coordinates{(1,0) (0.4333, 0) (0.4444, 0.0250) (0.8935, 0.1597)} --cycle;
                \addplot[name path = k234_2, color = black, fill = black, area legend] coordinates{(0.5417, 0) (0.9236, 0.1146) (1,0)} --cycle;

            \end{axis}
        \end{tikzpicture}%
    \end{lrbox}
    \resizebox{0.8\columnwidth}{!}{\usebox{\mybox}}
    \caption{Regions in the (normalised) $R_c-d_c$ space: \textit{Left}: for the Pythagorean triples: $\mathcal{S}_1=\{5,12,13\}$, $\mathcal{S}_2=\{8,15,17\}$, and $\mathcal{S}_3=\{12,35,37\}$; $k=2$ is fixed. \textit{Right}: for different values of $k$: $k_1=2,k_2=3,k_3=4$; the Pythagorean triple is fixed to $\mathcal{S}_1$. Distance parameters are the same as used to generate the regions in Fig.~\ref{fig:DiffRcDcEx}.}
    \label{fig:PyTripRcdc}
\end{figure}
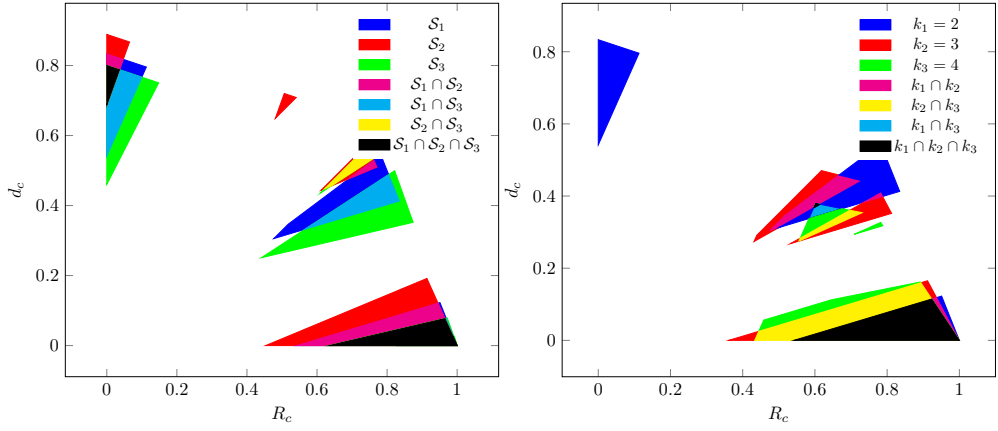

The results section is concluded by showing the influence of two design parameters: the Pythagorean triple and $k$, on the sizes of the feasible regions in the $R_c-d_c$ space. The change in these regions for three different triples, $\mathcal{S}_1=\{5,12,13\}$, $\mathcal{S}_2=\{8,15,17\}$, and $\mathcal{S}_3=\{12,35,37\}$, while keeping $k=2$ as a constant, is shown in the left sub-figure in Fig.~\ref{fig:PyTripRcdc}. Similarly, for $\mathcal{S}_1$ and varying $k$, the regions are shown in the right sub-figure in Fig.~\ref{fig:PyTripRcdc} To make comparison easier, the $R_c,d_c$ values for each set are normalised to lie between $[0,1]$. As can be seen, both parameters play a crucial role in the generation of feasible paths. While near circular paths can be generated by all sets, regions that lead to paths that resemble trochoids reduce considerably for some combinations. As discussed before, it helps to choose regions that are ``large'' in order to accommodate perturbations in initial positions and those lead to larger sensing area coverage.

\section{Conclusions}\label{Sec:Conc}

In this paper, a CP developed to enable the creation of artistic patterns is designed for a connected swarm of non-holonomic robots. By considering a swarm of 3 robots, parameters of closed and periodic trochoidal paths are calculated that ensure distance constraints are satisfied; these constraints consider robots of finite-size, hence guarantee collision avoidance if satisfied, and communication range between the robots. Further, constraints on speed and turn-rate as well as perturbations in the initial positions of the robots are also considered in the design. By exploiting the properties of the CP and the characteristics of trochoids, the proposed design procedure reduces the number of design variables and also allows for scaling temporal movements without causing any change in the geometric patterns. Conditions when additional robots may be added to the paths designed for 3 robots are also identified. The design of two types of closed paths - circular and epitrochoidal - is also implemented on a hardware platform; the trade-offs that result between these types of paths are also highlighted. 

This work can be naturally extended to the 3-D case, for instance, with ground robots replaced by drones, so that additional applications can be covered; another extension is considering more than 3 agents. Alternative solutions for the design of the CP can be considered that can yield non-closed paths, which may present desirable properties in some applications. Changing these design parameters dynamically, while ensuring the identified constraints continue to be satisfied, is another extension to this work.

\section*{APPENDIX}\label{Sec:App}

The coefficient vectors $\Gamma_R=[\gamma_{r1} \ \gamma_{r2} \ \gamma_{r3}], \cdots$ in \eqref{rcdcEqs} are given by
\begin{align*}
    \gamma_{r1} &= \beta_2\left(\beta_1 + 2\beta_2 + \beta_3 + 2b\right) \\
    \gamma_{r2} &= \beta_1\left(\beta_1 - \beta_3 + 2b\right) - 2\beta_2\beta_3 \\
    \gamma_{r3} &= -\left(\beta_1^2 + 2\beta_2^2 - \beta_1\beta_3 - \beta_2\beta_3 + \beta_1\beta_2 + 2b\beta_1 + 2b\beta_2\right) \\
    \gamma_{d1} &= \beta_2\left(\beta_1 + 2\beta_2 + \beta_3 - 2b\right)\\
    \gamma_{d2} &= \beta_1\left(\beta_1 - \beta_3 - 2b\right) - 2\beta_2\beta_3\\
    \gamma_{d3} &= -\left(\beta_1^2 + 2\beta_2^2 - \beta_1\beta_3 - \beta_2\beta_3 + \beta_1\beta_2 - 2b\beta_1 - 2b\beta_2\right)\\
    \gamma_{\phi r 1} &=  -\beta_2\left( \beta_1 + \beta_3 + 2\beta_2 + 2b\right) \\
    \gamma_{\phi r 2} &=  -\beta_1\left(\beta_1 - \beta_3 + 2b\right) + 2\beta_2\beta_3\\
    \gamma_{\phi r 3} &= \beta_1^2 + 2\beta_2^2 - \beta_1\beta_3 - \beta_2\beta_3 + \beta_1\beta_2 + 2b\beta_1 + 2b\beta_2 \\
    \gamma_{\phi d 1} &=  \beta_2\left(\beta_1 + \beta_3 + 2\beta_2 - 2b\right) \\
    \gamma_{\phi d2} &= \beta_1\left( \beta_1 - \beta_3 - 2b\right) - 2\beta_2\beta_3 \\
    \gamma_{\phi d 3} &= -\left(\beta_1^2 + 2\beta_2^2  - \beta_1\beta_3 - \beta_2\beta_3  + \beta_1\beta_2 - 2b\beta_1 - 2b\beta_2 \right)
\end{align*}

\bibliographystyle{unsrt}
\bibliography{TrochoidRef}  

\begin{thebibliography}{10}

\bibitem{Tsiotras2014}
Panagiotis Tsiotras and Luis~Ignacio Reyes~Castro.
\newblock {\em The Artistic Geometry of Consensus Protocols}, pages 129--153.
\newblock Springer International Publishing, Cham, 2014.

\bibitem{Fedele2023}
Giuseppe Fedele and Luigi D’Alfonso.
\newblock On the emergent hypocycloidal and epicycloidal formations in a swarm
  of double integrator agents.
\newblock {\em IEEE Control Systems Letters}, 7:613--618, 2023.

\bibitem{Fedele2023TAC}
Giuseppe Fedele, Luigi D'Alfonso, and Antonio Bono.
\newblock Emergent hypotrochoidal and epitrochoidal formations in a swarm of
  agents.
\newblock {\em IEEE Transactions on Automatic Control}, pages 1--8, 2023.

\bibitem{MONSINGH201984}
Jerome~Moses Monsingh and Arpita Sinha.
\newblock Trochoidal patterns generation using generalized consensus strategy
  for single-integrator kinematic agents.
\newblock {\em European Journal of Control}, 47:84--92, 2019.

\bibitem{M2023100928}
Jerome~Moses Monsingh, Arpita Sinha, and Hoam Chung.
\newblock Trochoidal patterns generation using generalized consensus strategy
  for double-integrator dynamic agents.
\newblock {\em European Journal of Control}, page 100928, 2023.

\bibitem{hegde2021}
Aditya Hegde and Anoop Jain.
\newblock Synchronization and balancing around simple closed polar curves with
  bounded trajectories and control saturation, 2021.
\newblock arXiv.2110.07248.

\bibitem{Pavone2007}
Marco Pavone and Emilio Frazzoli.
\newblock Decentralized policies for geometric pattern formation.
\newblock In {\em 2007 American Control Conference}, pages 3949--3954, 2007.

\bibitem{Borkar2020}
Aseem~Vivek Borkar, Arpita Sinha, Leena Vachhani, and Hemendra Arya.
\newblock {Application of Lissajous curves in trajectory planning of multiple
  agents}.
\newblock {\em Autonomous Robots}, 44(2):233--250, January 2020.

\bibitem{Boldrer2023}
Manuel Boldrer, Lorenzo Lyons, Luigi Palopoli, Daniele Fontanelli, and Laura
  Ferranti.
\newblock {Time-Inverted Kuramoto Model Meets Lissajous Curves: Multi-Robot
  Persistent Monitoring and Target Detection}.
\newblock {\em IEEE Robotics and Automation Letters}, 8(1):240--247, 2023.

\bibitem{Ratnoo2019}
P~Anjaly and Ashwini Ratnoo.
\newblock {\em Analysis of Range-based Agent Motion Patterns}.

\bibitem{Parayil2019}
Anjaly Parayil and Ashwini Ratnoo.
\newblock Bifurcation-based control law for pattern generation.
\newblock {\em IEEE Control Systems Letters}, 3(2):374--379, 2019.

\bibitem{Keller2017}
James Keller, Dinesh Thakur, Maxim Likhachev, Jean Gallier, and Vijay Kumar.
\newblock {Coordinated Path Planning for Fixed-Wing UAS Conducting Persistent
  Surveillance Missions}.
\newblock {\em IEEE Transactions on Automation Science and Engineering},
  14(1):17--24, 2017.

\end{thebibliography}

\end{document}